\documentclass[]{gtech}
\usepackage{amssymb}
\usepackage{multirow}
\usepackage{bigdelim}
\usepackage{todonotes}
\usepackage{longtable}
\usepackage{tabularray}
\usepackage{wrapfig}
\usepackage[most]{tcolorbox} 
\usepackage{xcolor}
\usepackage[table]{xcolor}
\usepackage{url}
\usepackage{listings}

\renewcommand{\title}[1]{\newcommand{\titlelist}{{\huge\fontfamily{optimistic}\selectfont #1}}}

\newcommand{\model}{\texttt{Ming-Omni}}
\newcommand{\modellite}{\texttt{Ming-Lite-Omni}}
\newcommand{\modelflash}{\texttt{Ming-Flash-Omni}}

\definecolor{prompt}{HTML}{5f84e4}
\definecolor{img}{HTML}{820100}

\usepackage{arydshln}

\definecolor{CQColor}{rgb}{0.0,0.0,1.0} % color for Aaron

\definecolor{TSColor}{rgb}{0.5,0.0,0.8} % color for Aaron

\definecolor{CQRColor}{rgb}{1.0,0.0,1.0} % color for Aaron

\usepackage{colortbl}
\usepackage{amssymb}
\usepackage{pifont}
\usepackage{booktabs,multirow}
\usepackage{makecell}
\usepackage{tabulary}
\usepackage{fontawesome5}
\usepackage{bbding}
\usepackage{multicol}

\newlength\savewidth

\title{\textcolor[HTML]{0369ff}{Ming-Flash}-Omni: A Sparse, Uni{f}{i}ed Architecture for Multimodal Perception and Generation}
\author[*]{Inclusion AI, Ant Group}

\contribution[*]{See Contributions section (Sec.~\ref{sec:contri}) for full author list.}

\abstract{\fontsize{11pt}{12pt} \textit{We propose Ming-Flash-Omni, an upgraded version of Ming-Omni, built upon a sparser Mixture-of-Experts (MoE) variant of Ling-Flash-2.0 with 100 billion total parameters, of which only 6.1 billion are active per token. This architecture enables \textbf{highly efficient scaling} (dramatically improving computational efficiency while significantly expanding model capacity) and empowers stronger unified multimodal intelligence across vision, speech, and language, representing a key step toward Artificial General Intelligence (AGI). Compared to its predecessor, the upgraded version exhibits substantial improvements across multimodal understanding and generation. Notably, it achieves strong performance on \textbf{vision-language understanding benchmarks}, with overall scores \textbf{on par with Gemini 2.5 Pro}, and enables \textbf{seamless switching among multimodal tasks} in multi-turn interactions. In speech, it achieves strong performance in \textbf{contextual and dialect-aware ASR} while enabling \textbf{joint, continuous-generation of speech, sound, and music}. In vision, it introduces \textbf{generative semantic segmentation} that achieves competitive standalone performance and enhances spatial control and editing consistency, alongside marked improvements in \textbf{identity preservation}, and \textbf{high-fidelity in-image text rendering}. Together, these capabilities demonstrate that a single unified model can serve as a practical foundation for general-purpose multimodal intelligence.%Notably, Ming-Flash-Omni achieves state-of-the-art results in text-to-image generation and generative segmentation,  all within a single unified architecture. 
}}

\date{Mar 26, 2026\vspace{-1mm}}
\gtechdata[Project Homepage]{\url{https://github.com/inclusionAI/Ming}}
\begin{document}
\maketitle

\section{Introduction}
\label{sec:intro}

% 开头背景 
% 结构上关键升级 
% 能力提升 & 新增能力 
% 具体值得highlight的指标

% \label{sec:method}
\begin{figure}[t]
    \centering
    \includegraphics[width=1.0\linewidth]{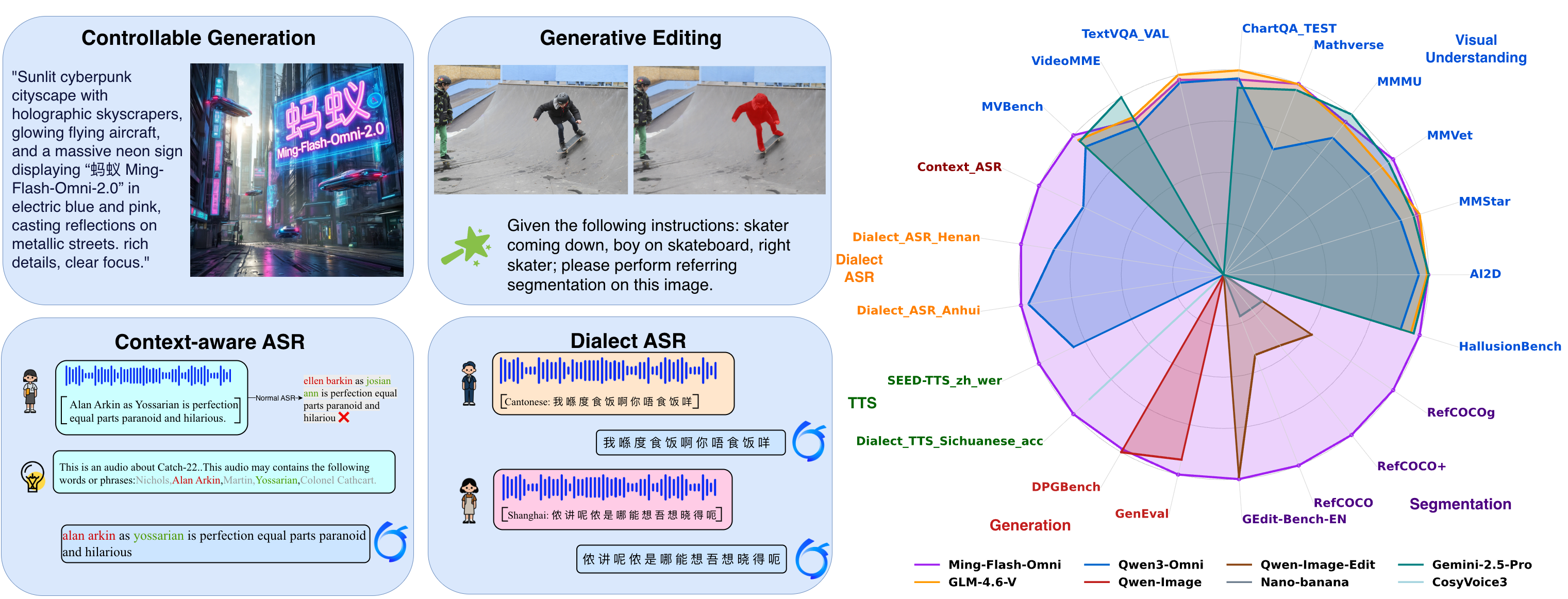} 
    \caption{\modelflash{} generally demonstrates highly competitive performance across various domains, including vision-text understanding, controllable image generation, speech recognition, and speech synthesis. Specifically, in image generation, \modelflash{} introduces a novel generative segmentation paradigm to achieve fine-grained spatial and semantic control over the generated images. Moreover, \modelflash{} significantly enhances Context-Aware Speech Recognition (ContextASR) and Chinese dialect recognition, thereby broadening its applicability in real-world scenarios.}
    \label{Ming_moe_uni_example}
    % \vspace{-4mm}
\end{figure}

In everyday life, humans naturally integrate visual and auditory cues to express ideas through speech or writing, while also forming vivid mental images from descriptions or concepts. This ability to visualize enhances creativity, problem-solving, and communication, serving as a core aspect of human intelligence and interaction.
The ultimate goal of Artificial General Intelligence (AGI) is to replicate this human-like multimodal intelligence, evolving from a mere tool into a powerful agent that augments and liberates human productivity.

Driven by advances in Large Language Models (LLMs) and extensive training on large-scale multimodal datasets, Multi-modal Large Language Models (MLLMs) have demonstrated remarkable perceptual capabilities in both vision~\citep{chen2024internvl,bai2025qwen2,team2025kimi,xu2025qwen3} and audio~\citep{ding2025kimi,xu2025qwen2,xu2025qwen3}, as well as generative capabilities in these two modalities~\citep{huang2025step,ding2025kimi,chatgpt4o,tong2024metamorph,MingUniVision2025,pan2025transfer,xu2025qwen3}. Nevertheless, effectively integrating comprehension and generation across multiple modalities into a unified model remains challenging. While humans naturally learn by combining multiple modalities, leveraging their complementary strengths and interactions to enhance overall learning efficiency, building a unified Omni-MLLM is hindered by representational disparities and modality imbalances.

In this paper, we introduce \modelflash{}, which builds upon the \model{} architecture with a redesigned foundation and targeted enhancements across multimodal understanding and generation. At its core, \modelflash{} adopts Ling-Flash-2.0~\cite{lingv2_flash} (a scaled-up, highly sparse Mixture-of-Experts architecture) where an increased sparsity ratio enables substantial model capacity while maintaining bounded inference latency, striking a favorable trade-off between performance and efficiency. 

On the understanding side, the model introduces three key advances. First, \modelflash{} upgrades the positional encoding to time-interleaved VideoRoPE~\cite{wei2025videorope}, which interleaves timestamps with video frames to align each visual token with its precise temporal position, better capturing video temporal dynamics. Second, ~\modelflash{} focus on improving the context-aware ASR capability itself, enhancing the model’s ability to leverage surrounding linguistic context during speech recognition and thereby achieving more accurate transcription in context-dependent scenarios. Third, the model supports multi-turn, cross-task understanding, enabling seamless switching between heterogeneous comprehension tasks (\textit{e.g.}, visual QA, audio captioning, document analysis) within a single dialogue session.

On the generation side, \modelflash{} introduces three key advancements: 1) beyond zero-shot voice cloning, it enables joint, single-channel generation of speech, sound, and music with fine-grained vocal control, replacing discrete acoustic tokens with continuous representations to circumvent quantization artifacts and produce more natural, expressive TTS outputs; 2) the model supports generative semantic segmentation, enabling pixel-level semantic content generation conditioned on multimodal inputs; and 3) it enables fine-grained controllable image generation with improved identity preservation, high-fidelity in-image text rendering, and robustness enhanced through vision-specific reinforcement learning.

These architectural innovations empower \modelflash{} to deliver exceptional cross-modal performance in both comprehension and generation tasks. Specifically, in image perception, \modelflash{} attains performance comparable to that of Gemini 2.5 Pro~\citep{comanici2025gemini25}; in image generation, it supports generative semantic segmentation and fine-grained controllable synthesis with strong identity preservation and high-fidelity in-image text rendering; and in speech, it achieves state-of-the-art end-to-end understanding and generation capabilities.

The remainder of this paper is organized as follows. Section 2 presents the detailed architecture of \modelflash{}. Sections 3 describes the pretraining and post-training datasets. 
Section 4 reports the evaluation results and compare \modelflash{} with recent multimodal models. Sections 5 is conclusion.

%The key features of \model{} can be summarized as follows.
%\begin{itemize}

%\item \textbf{Unified Omni-Modality Perception:}
% \model{}, based on Ling (an MoE architecture LLM), resolves task conflicts and ensures harmonious integration of tokens from different modalities through modality-specific routers.
%\model{}, built on Ling~\citep{ling}, an MoE architecture LLM, resolves task conflicts and ensures coherent integration of tokens from different modalities through modality-specific routers.

%\item \textbf{Unified Perception and Generation:}
%\model{} achieves unified understanding and generation, enabling the model to interpret multimodal instructions and user intent during generation, thereby improving generation quality and usability across multiple tasks.

%\item \textbf{Innovative Generation Capabilities:} %\%model{} can perceive all modalities and generate high-quality text, real-time speech, and vivid images simultaneously, delivering exceptional cross-modal performance across diverse tasks including image perception, audio-visual interaction, and image generation.
%\end{itemize}

% \clearpage

\section{~\modelflash{}}
\label{sec:method}

\begin{figure}[t]
    \centering
    \includegraphics[width=0.85\linewidth]{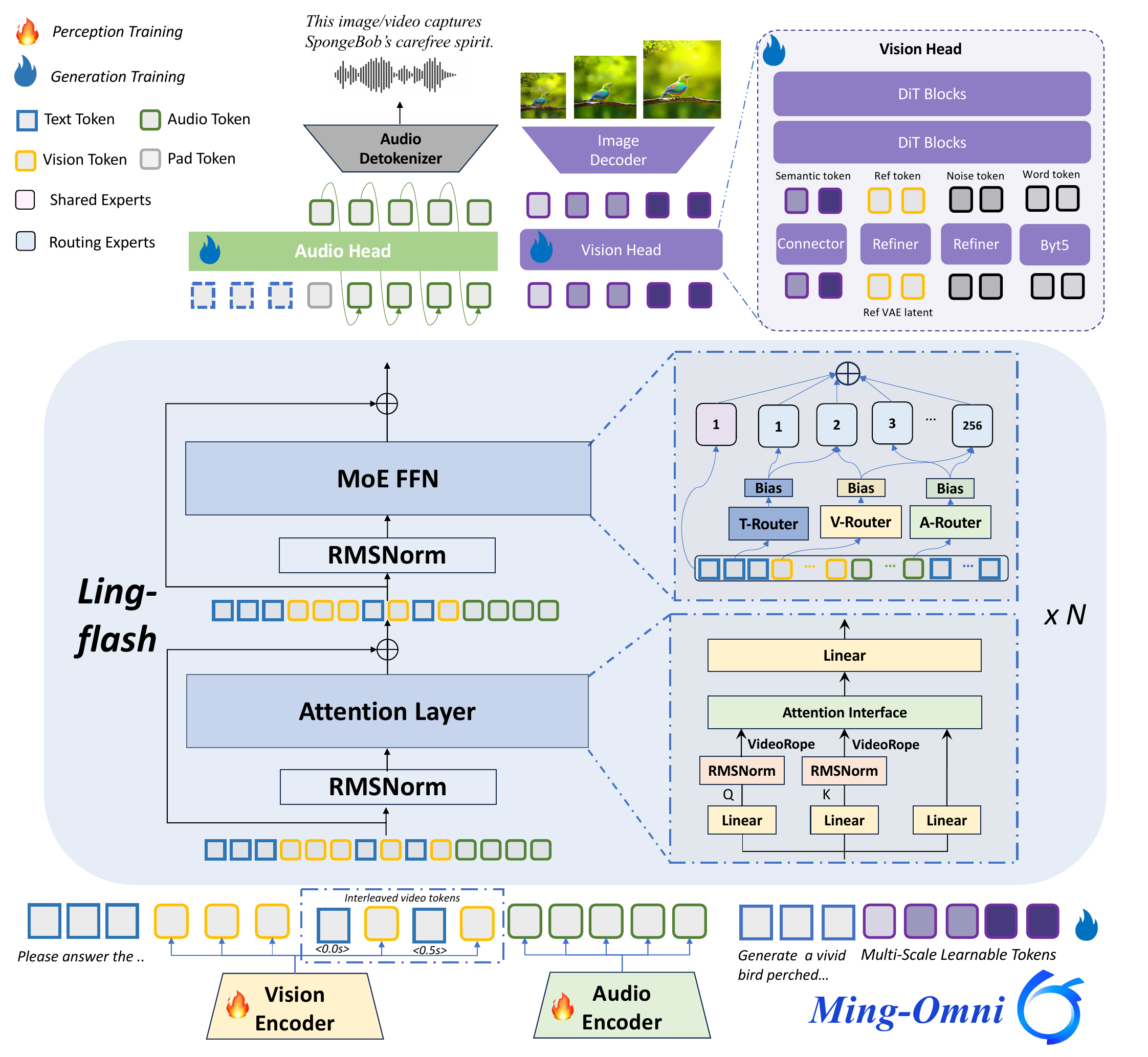}
    \vspace{-2mm}
    \caption{The overall framework of \modelflash{}. This version features a sparser LLM based on Ling-flash-2.0 MoE architecture, and integrates VideoRoPE to enhance temporal modeling. Speech generation now uses continuous features instead of discrete tokens, and image generation has been upgraded with support for segmentation.}
    \label{Ming_moe_uni}
    \vspace{-2mm}
\end{figure}

As illustrated in Figure~\ref{Ming_moe_uni}, \modelflash{} retains the unified two-stage pipeline of \model{}~\cite{ai2025ming}, where perception supports multimodal understanding and generation targets speech and image synthesis, while markedly advancing long-context modeling, reasoning, and controllable generation. At the core is Ling‑Flash‑2.0~\cite{lingv2_flash}, a sparse MoE LM (100B; 6.1B per token) with a dual balancing scheme that stabilizes training and improves efficiency. On the perception side, \modelflash{} employs time-interleaved VideoRoPE, an enhanced variant of VideoRoPE~\cite{wei2025videorope} that interleaves timestamps with video frames to achieve fine-grained temporal alignment, explicitly associating each visual token with its precise temporal position; it also integrates context‑aware ASR for more reliable speech understanding and a think mode for deeper multi‑step reasoning. On the generation side, we replace discrete speech tokens with continuous acoustic latents, avoiding quantization loss and improving fidelity; for images, we upgrade to a synergistic training paradigm that enables generative segmentation-as-editing, facilitating fine-grained and controllable generation. Overall, \modelflash{} advances the unified model with stable expert routing and scalable long‑context modeling, yielding more reliable multimodal understanding and controllable generation.

\subsection{Unified Understanding Across Modalities}

A core challenge in unified multimodal models is the effective fusion of image understanding and generation. While our \modelflash{} injects hierarchical semantics via multi-scale query tokens, its language pathway remains frozen during training to prevent interference from the generative objective. Although this ensures stability, it creates a critical bottleneck: a misalignment between understanding and generation objectives. Consequently, even with injected semantics, fine-grained visual knowledge (such as object attributes and spatial relationships) cannot be efficiently transferred to high-precision generation and editing, limiting final quality and controllability.

%The cornerstone of \modelflash{} is enhanced multimodal understanding. We retain the established visual and audio encoders (Qwen3-VL~\citep{Qwen3-VL} and Whisper~\citep{radford2023robust}) and feed their projected embeddings, concatenated with tokenized text, into Ling‑flash‑2.0, a sparse MoE language model with distinct routers per modality. Beyond this, \modelflash{} incorporates VideoRoPE to maintain temporal coherence over long-range frame sequences, thus emphasizing temporal modeling. Furthermore, \modelflash{} adopts a context‑aware ASR training paradigm that conditions decoding on task or domain context (leveraging preceding textual context and hotword lists during training) to address common shortcomings of conventional ASR in real‑world, multi‑domain scenarios, such as limited world knowledge and unreliable proper‑noun recognition, thereby yielding more accurate and context‑consistent transcripts. To stabilize training in the more sparse Ling‑flash‑2.0, we employ a hybrid expert‑balancing scheme that combines an auxiliary load‑balancing loss (as in \model{}) with per‑router bias updates (as in Ling‑flash‑2.0), promoting uniform expert activation and improved convergence. 

\subsection{Unified Speech Understanding and Generation}
\label{sec_2_speech_understanding}

\modelflash{} unifies speech understanding and generation within a single architecture. For generation, we replace discrete speech tokens with continuous acoustic latents, avoiding quantization loss and improving fidelity. Specifically, we utilize a fixed, pre-trained audio head based on Qwen2.5 (0.5B parameters), which takes LLM-generated text tokens and downsampled VAE latents as input and autoregressively predicts the conditioning signals for the flow-matching head—following the paradigm of~\cite{jia2025ditar}. 

% Beyond zero-shot voice cloning, ~\modelflash{} enables joint, single-channel generation of speech, sound, and music, along with fine-grained control over vocal attributes. This is powered by a novel VAE-based audio tokenizer operating at 12.5 Hz, which establishes a unified acoustic latent space spanning speech, music, and environmental sounds. Controllability is achieved through a multi-stage training curriculum that implicitly disentangles timbre from textual instructions,  enabling voice design via natural language descriptions. On the Instruct-TTS-Eval-zh benchmark~\citep{huang2025instructttseval}, \modelflash{} matches the performance of Qwen3-TTS~\citep{hu2026qwen3}. Additionally, the model incorporates robust text normalization to accurately render complex expressions such as mathematical and chemical equations.

Beyond zero-shot voice cloning, \modelflash{} enables joint, single-channel generation of speech, sound effects, and music, alongside fine-grained control over vocal attributes. This capability is underpinned by a novel variational autoencoder VAE-based audio tokenizer. To address two key challenges, cumulative error in autoregressive (AR) modeling caused by the inherently long sequence lengths of raw audio, and the mismatch in sampling rates between high-fidelity music (typically 44.1 kHz) and speech, we introduce a low-frame-rate tokenizer operating at 12.5 Hz. This design substantially compresses the token sequence length, thereby mitigating error propagation during AR generation. Moreover, we unify all training data to a consistent 44.1 kHz sampling rate and integrate a built-in super-resolution module, ensuring seamless compatibility with inputs originally sampled at lower rates.

Regarding controllability, in zero-shot scenarios, reference audio often contains extraneous attributes beyond timbre, such as emotional prosody or speaking style, that can interfere with instruction-guided generation. Conventional approaches typically address this by employing specialized components like gradient reversal layers to disentangle timbre from other acoustic factors; however, such methods introduce considerable architectural and training complexity. To address this, we designed a multi-stage training strategy: the first stage focuses on learning fundamental text-to-audio mappings, while the second stage concurrently injects voiceprint features extracted from reference audio alongside textual instructions. Given the robust pronunciation foundation established in the first stage, the model rapidly learns to map voiceprints to timbre and text instructions to style attributes, ultimately achieving disentanglement of timbre and control instructions without requiring any additional modules. On the Instruct-TTS-Eval-zh benchmark~\citep{huang2025instructttseval}, \modelflash{} matches the performance of Qwen3-TTS~\citep{hu2026qwen3}. Moreover, the model incorporates a robust text normalization pipeline capable of accurately rendering complex symbolic expressions, including mathematical equations and chemical formulas.

\subsection{Unified Image Understanding and Generation}
A core challenge in unified multimodal models is the effective fusion of image understanding and generation. While our \model{} injects hierarchical semantics via multi-scale query tokens, its language pathway remains frozen during training to prevent interference from the generative objective. Although this ensures stability, it creates a critical bottleneck: a misalignment between understanding and generation objectives. Consequently, even with injected semantics, fine-grained visual knowledge (such as object attributes and spatial relationships) cannot be efficiently transferred to high-precision generation and editing, limiting final quality and controllability.

We address this through three synergistic pillars: 
1) a single-stream diffusion transformer that unifies cross-modal representation; 
2) a generative segmentation pretraining task that tightly couples perception and generation; 
and 3) a robust multi-reward reinforcement learning framework that refines controllability without overfitting.

\paragraph{\textbf{Single-Stream Diffusion Transformers}} 

To overcome semantic fragmentation in dual-stream architectures (\textit{e.g.}, MMDiT), we adopt a native single-stream diffusion transformer~\citep{cai2025z}. In dual-stream architectures, the noisy latent and reference latent are processed with independent self-attention, which exacerbates copy‑and‑paste artifacts from the reference image. In single-stream architectures, text, images, and other modalities are mapped into a unified token space, with full cross-modal attention in a single stream—breaking modality boundaries. Unlike the “multi-image collage” artifacts of prior methods, our approach achieves global semantic coherence in lighting, proportions, and layout, drastically reducing visible “copy-paste” effects.

\paragraph{\textbf{Generative Segmentation as an Editing Task}} 
Building on this unified representation, we bridge the understanding–generation gap by reframing image segmentation as a generative editing task. Instead of producing abstract binary masks (\textit{e.g.}, ``segment the banana''), the model performs semantics-preserving edit (\textit{e.g.}, ``color the banana purple''). This reformulation tightly couples understanding and generation: accurate editing requires precise perception of object boundaries, making comprehension a \textit{prerequisite} for generation. The edit quality thus provides direct supervision for visual understanding, unifying their optimization objectives. As a result, the model acquires fine-grained spatio-semantic control, which crucially addresses compositionality challenges in text-to-image generation.

\paragraph{\textbf{Robust vision generation reinforcement learning}}
To further refine alignment and controllability, we adopt a post-training RL framework that avoids SDE solvers and accelerates training by optimizing over a subset of diffusion steps~\citep{zheng2025diffusionnft}. To mitigate reward hacking, we initialize with a generative segmentation task and replace the KL divergence constraint with offline data regularization. We then perform sequential RL training on image generation and editing, using multi-dimensional rewards (realism, instruction adherence, aesthetic quality, and task-specific scores) to prevent overfitting to any single metric.

\paragraph{\textbf{Empowering Advanced Controllable Capabilities}}
The synergy of these components unlocks a suite of advanced, high-fidelity controllable functions:

\textit{Identity (ID) Preservation} To preserve the identity of the reference subject, we concatenate the VAE-encoded features of the reference image with the noisy latent representation of the diffusion model along the token dimension and enhance identity consistency through self-attention mechanisms. This approach is further complemented by learnable query tokens that encode global semantic information from the reference image. Together, these two complementary information streams ensure structural and appearance consistency in unedited regions while simultaneously improving the fidelity and accuracy of the edited content. For highly dynamic motion scenes (such as those commonly encountered in travel photography), we introduce motion-aware action label representations. By employing a balanced sampling strategy across these action labels during training, our model learns to effectively generate and edit motion-related regions. This strategy significantly mitigates the occurrence of unnatural poses and enhances the overall realism of the synthesized results.

\textit{High-Fidelity Text Rendering.} By integrating a specialized Glyph-byT5 text encoder, our model leverages its learned pixel-level control to accurately place text, ensuring seamless contextual integration and high-quality results. Specifically, we feed the text to be rendered into a ByT5 encoder to obtain glyph features, which are then used together with learnable query tokens as conditioning inputs to the diffusion model. The advantage of this approach is faster training convergence, while the downside is that, during inference, we must first extract the text to be rendered from the user’s prompt in advance.

\subsection{Overall Training Procedure}

The training procedure of \modelflash{} retains a two-stage pipeline: perception and generation. The perception stage of \modelflash{} comprises three sequential phases: progressive pre-training, two-stage instruction tuning, and reinforcement learning. Pre-training spans three stages: two at 8K context length to build robust multimodal foundations, followed by a 24K stage to capture long-range cross-modal dependencies. Instruction tuning then proceeds in two phases: Stage I uses 10K sequences over text, images, and audio for basic instruction following; Stage II extends the context to 64K, integrating video and multi-turn dialogue to enable temporally coherent multimodal interaction. During the reinforcement learning stage, we adopt \emph{Group Relative Policy Optimization} (GRPO)~\cite{shao2024deepseekmath} to train the model. To accommodate different task types, we construct a domain-specific reward system that integrates rule-based rewards, model-based rewards, and checklist-based general rewards, enabling the unified reinforcement learning of heterogeneous tasks within a single training stage. To further improve the model’s continuous interaction capability across arbitrary modalities, we introduce multi-turn modality-switching conversation optimization over image, audio, video, and text. 

After perception, we freeze the perception MLLM and optimize only the image generator, while leveraging the pre-trained audio generator from ~\cite{MingUniAudio2025}. For image generation, the training procedure contains three sequential stages. In the first stage, we pre-train a diffusion-based image generator using a flow matching objective, while keeping the perception MLLM frozen. The generator is equipped with multi-scale learnable queries to capture hierarchical visual semantics from textual inputs. In the second stage, we extend the model to support image editing by conditioning the denoising process on reference images: the VAE-encoded representations of input images are concatenated with the noisy latent to enforce structural and semantic consistency with the original content. Additionally, input word-level captions are encoded via ByteT5 embeddings to enrich textual conditioning. In the third stage, we conduct RL training sequentially on image generation and image editing, integrating multi-dimensional objectives, including image realism, instruction adherence, aesthetic quality, and task-specific rewards.

\subsection{Infrastructure}

Compared to large language models (LLMs), the training of multimodal foundation models is characterized by distinctive challenges rooted in \textbf{data heterogeneity} and \textbf{model heterogeneity}. First, data heterogeneity arises from the need to dynamically switch between diverse input modalities (text, images, audio, and video) during training. These modalities exhibit significant differences in tensor shape, most notably in the form of dynamic batch sizes and variable-length sequences. This variability complicates the design of a unified parallel computation layout. As a result, computational workloads become unevenly distributed across processing ranks, leading to load imbalance. Moreover, the frequent allocation and deallocation of GPU memory buffers for inputs of varying shapes induce severe memory fragmentation, substantially degrading training efficiency and hardware utilization. Second, in contrast to large language models (LLMs), which are predominantly based on homogeneous, decoder-only Transformer architectures, multimodal foundation models typically employ modality-specific encoders at the input stage, introducing model heterogeneity. Despite their modest size, these encoders are adversely affected by parallelization strategies that introduce substantial pipeline bubbles (\textit{i.e.}, idle computation cycles). This inefficiency becomes a major bottleneck for training throughput.

To address these challenges, ~\modelflash{} is trained on an enhanced version of the Megatron-LM~\cite{shoeybi2019megatron} framework with three key extensions tailored for multimodal workloads:

\begin{itemize}
    \item {Sequence Packing For Data Heterogeneity:} To support dynamic input shapes, we adopt sequence packing, which packs multiple variable-length samples into fixed-length batches. In addition, to ensure data consistency across PP stages, we design a communication protocol that overlaps communication with computation. Together, these methods improve memory utilization and computational density.
    \item {Flexible Parallelization for Model Heterogeneity:} To address workload imbalance caused by architectural asymmetry between encoder and decoder, we extend Megatron-LM to support flexible parallelization, enabling independent DP/PP/TP configurations for each component. Furthermore, we formulate the pipeline dependency as a linear programming problem to automatically derive optimal stage layouts for varying sequence lengths. These synergistic optimizations yield end-to-end training speedups of 58.1\%–69.7\% over uniform manual setups across multiple model scales. 
\end{itemize}

Additionally, we observe that state-of-the-art memory-intensive operators (especially during backward passes) exhibit low memory bandwidth utilization, achieving only a fraction of the theoretical hardware peak. To address this inefficiency, we design custom CUDA kernels (\textit{e.g.}, for LayerNorm and RMSNorm) that raise bandwidth utilization to approximately 80\% of the hardware limit. Our optimized kernels reduce forward and backward latency by up to 25.3\% compared to Transformer Engine and deliver a 5× speedup over native PyTorch.

Collectively, these optimizations enable a Model FLOPs Utilization (MFU) of 28\%, representing more than a 4x improvement  in training throughput over the baseline Megatron-LM implementation.

% \input{sections/3_app.tex}
% \newpage

\section{Data Construction}
\label{sec:data}

% ------------------------- section -------------------------------

We have collected a large and diverse set of training data to enable models to process and understand information from multiple modalities, including text, images, audio and videos.
The majority of this data comes from \model{}~\citep{ai2025ming}. In addition, we develop several data processing pipelines to ensure data quality, diversity and deduplication. Establishing an effective multimodal data strategy is essential for the joint multi-modal training, as it facilitates seamless alignment of knowledge across diverse modalities.
We categorize the training data based on the core modalities they are designed to enhance, including image, audio, video, and text. The detailed sources and construction methods for each type of data are elaborated in this section.

% 划分依据，video centric，数据分类依据，核心目标是为了优化哪个模态 ok

\subsection{Image Data}
Image data serves as the cornerstone of our multi-modal corpus.
Following \model{}~\citep{ai2025ming}, we integrate both image-understanding and image-generation datasets to enable the MLLM to acquire unified perception and generation capabilities. Additionally, we further design novel pipelines to synthesize high-quality datasets across diverse dimensions to improve model's capabilities and user interaction quality.
%data quality while also reducing the data volume.

\subsubsection{Image Understanding Data}

\textbf{Knowledge Data:} To enhance \modelflash{}’s expert-level comprehension and perception in knowledge-intensive tasks, we construct a large-scale, high-fidelity encyclopedia data ecosystem and directly integrate it into \modelflash{}’s training pipeline. This integration enables \modelflash{} to perform expert-level reasoning, such as accurately identifying rare or endangered species using Latin binomial nomenclature. The corpus encompasses a broad range of knowledge domains, with representative coverage across three thematic categories: biological (Animals, Plants), cultural (Celebrities, Anime Characters, Landmarks), and daily-life (Dishes, Antiques, Calligraphy and Paintings). Data are sourced from academic databases, institutional websites, digital museums, and specialized professional platforms. 

We design two complementary data pipelines to generate synergistic forms of knowledge-intensive training data, both directly used to train \modelflash{} for multimodal alignment and reasoning:

\begin{itemize}
    \item {Expert-Level Entity Recognition Data:} Using a “breadth-to-depth” strategy, we harvest canonical entities (\textit{e.g.}, species binomials) and retrieve semantically relevant images via search engines. A progressive filter (combining CLIP consistency, MLLM-based verification, and manual refinement) ensures high precision, enabling \modelflash{} to recognize fine-grained expert concepts.
    \item {Encyclopedia-Based Knowledge QA:} We automatically align visual entities with structured knowledge by extracting ⟨image, entity, knowledge⟩ triplets, filtering them via a multi-VLM consensus system for cross-modal consistency, and converting validated triplets into ⟨image, question, answer⟩ VQA pairs using an LLM. This yields a broad-coverage dataset that significantly strengthens \modelflash{}’s knowledge-grounded reasoning.
\end{itemize}

\textbf{STEM Reasoning Data:}
% 推理数据
In the reasoning training of \modelflash{}, we enrich chain-of-thought (CoT) data to strengthen the model’s capacity for complex, multi-step STEM reasoning. The curated reasoning dataset is centered on two core themes: multidisciplinary reasoning and mathematical reasoning. To ensure high quality and structural coherence, we develop a systematic pipeline for CoT generation and filtering, comprising four key stages: (1) \textit{QA extraction}: We extract raw question–answer pairs from diverse sources; (2) \textit{Difficulty-aware filtering}: We discard samples deemed too simple for \modelflash{} based on preliminary model evaluation; (3) \textit{CoT expansion}: We leverage state-of-the-art multimodal reasoning models ~\citep{vteam2025glm45vglm41vthinkingversatilemultimodal,Qwen3-VL,GEMINI} to generate detailed, step-by-step reasoning traces, forming an initial CoT pool; (4) \textit{Quality validation}: We rigorously assess the logical consistency and factual correctness of the synthesized CoT responses and filter out low-quality or erroneous examples. This pipeline yields 1.6 million multimodal long CoT samples, with individual sequences reaching up to 8K tokens. Empirical results show that this data significantly enhances \modelflash{}’s performance on challenging STEM reasoning benchmarks.

\textbf{OCR Data:}
Text recognition and document understanding capabilities are crucial for MLLM. We construct a large-scale heterogeneous training dataset with millions of samples, consisting of three data sources: open-source data, expert-collaborative pseudo-labeled data, and human-annotated enhancement data. The expert-collaborative pseudo-labeled data is generated by diagnosing model weaknesses, and using expert models to label targeted data.
In addition, to enhance the model’s capability in text–visual analysis and logical reasoning, we incorporate the Chain-of-Thought (CoT) paradigm into the training data. We incorporate the open-source ChartQA-PoT dataset to enhance the model’s numerical reasoning ability on charts and pioneeringly use executable Python code as the intermediate reasoning representation.

\textbf{Safety Data:} We treat safety as an intrinsic capability of intelligent systems, aligning model behavior with ethical standards while preserving naturalness and helpfulness. This is achieved through a context-aware synthetic pipeline that expands unimodal safety seeds into a multimodal corpus at nearly 30× scale, structured along two axes: (1) illegal content and conduct (\textit{e.g.}, drugs, violence, pornography) and (2) socially sensitive topics governed by policy norms (\textit{e.g.}, politics, financial ethics). For query construction, we use multimodal large language models (MLLMs) to extract context-specific keywords from seed examples to guide image retrieval, then pair the retrieved images with synthesized prompts encoding harmful intents, adversarial metaphors, or high-risk inquiries. For response generation, we adopt a dual-strategy framework: queries involving strictly prohibited content elicit explicit refusals accompanied by corrective guidance, while safe but sensitive inquiries receive informative, boundary-respecting responses that maintain helpfulness without compromising safety.

\textbf{GRPO Data:} During the reinforcement learning (RL) stage, we train the model on multi-task data spanning a broad spectrum of general-purpose application scenarios, including knowledge-based question answering, mathematical and logical reasoning, multimodal instruction following, and OCR. The data is sourced from diverse tasks in the instruction tuning phase (\textit{e.g.}, conversational QA and multimodal understanding) and further augmented with user-experience-oriented examples (emphasizing fluency and practicality) as well as safety-critical cases (covering harmful content avoidance, privacy preservation, and compliance constraints). To ensure quality and diversity, we construct the RL dataset using task-type proportion control, difficulty-level stratification, and rigorous data cleaning and resampling strategies, yielding a curriculum specifically designed to strengthen the model’s general-purpose capabilities.

\subsubsection{Image Generation Data}

% 生产式分割数据，libin/lixiang补充相关数据

Image generation data enhances MLLM capabilities beyond traditional image understanding tasks. Building on the image generation data used in \model{}~\citep{ai2025ming}, we further incorporate segmentation, text rendering, and portrait preservation data to improve user experience. We extract image–text embeddings, apply k-means clustering, and use the resulting cluster labels to construct a semantically balanced training set via cluster-aware sampling. 

\textbf{Image segmentation data:} To improve the model’s generative segmentation capability, we construct two types of data: (1) We use the open-source referring segmentation datasets RefCOCO/+/g~\citep{refcoco,refcocog} to construct image editing data. The original image serves as the reference, and binary masks highlight target regions with specified colors to create edited images. (2) For semantic and panoptic segmentation, samples are built from COCO-Panoptic~\citep{Eval_mscoco,kirillov2019panoptic}, where each class or instance is assigned a unique color via a predefined colormap to generate edited images.

\textbf{Portrait preservation data:} The portrait preservation data consist of two data sources: (1) ID Photo Dataset: We collected and constructed 200k paired lifestyle-ID photos. And filter the data using four criteria, \emph{e.g.}, face similarity, face size and confidence, face angle and manual review. 
(2) Landmark Check-In Portrait Dataset: We collect 20K high-quality portraits from 225 landmarks.% (64 domestic and 161 international). 
The original images serve as edited images, while using advanced segmentation model like SAM2~\citep{ravi2024sam} to segments the foreground person and places them onto 1,000 manually collected daily background scenes as pre-edited images. And then use LLM generate diverse prompts for each landmark.

% Text generation data如何生成的prompt加到补充材料
\textbf{Text generation data:} 
We build a Chinese-English text generation dataset across three difficulty levels:
(1) Monotonic background text rendering: Text is rendered directly by setting background color, font type, size, color, and position.  (2) Text rendering on existing images: texts are rendered on suitable smooth regions obtained by Felzenszwalb algorithm. (3) Text-image integrated rendering: Using the SOTA LLMs %(Qwen2.5-32B~\citep{team2024qwen2} and ) 
to generate text rendering prompts, the advanced generation models (\emph{e.g.}, Qwen-image~\citep{wu2025qwen} and Nano Banana) are used for image generation, followed by OCR for consistency checks, resulting in a high-quality dataset.

% tts相关的训练数据，highlight模型tts效果可以，剑游 ok
\subsection{Audio Data}
For audio data, we mainly use the data from \model{}~\citep{ai2025ming}. In addition, we incorporate the following three datasets to further enhance the model’s audio understanding and generation capabilities.

\textbf{Context ASR Data:} 
Current ASR systems face challenges in recognizing homophones or phonetically similar words when the context is limited, pronunciations are unclear, or accents are present. ContextASR addresses these issues by leveraging the preceding context. We propose to synthesize a large-scale dataset using LLMs to endow models with ContextASR capabilities.
We extract named entities and construct context passages using LLMs based on existing ASR text, producing 3 million Chinese and English samples in the format <audio, text, context, entity\_list>. During training, we further filter and sample the data to reduce keyword density, remove keywords that are absent from the text, and generate negative samples to enhance the model’s discriminative ability. %The format of the training data is shown in Figure \ref{audio_sample}.

\textbf{TTS Data:} 
The diversity of TTS data is essential for fully leveraging the pretrained language model’s capabilities in audio generation. In addition to the open-source data used in \model{}, we develop a data generation pipeline to create large-scale TTS data. Specifically, (1) we crawl extensive audio data using keywords expanded from handcrafted seeds through domain-specific lexical variations, (2) apply VAD~\citep{gao2023funasrfundamentalendtoendspeech} to segment well-conditioned short clips, and (3) iteratively train an audio labeler—initially on high-quality data, then using its predictions to annotate the corpus with fine-grained labels for dimensions such as age, personality, style, speed, pitch, profession, emotion, and dialect. We curated a large-scale dataset of high-quality audio clips from diverse domains. These clips are annotated with a rich set of multi-dimensional labels, which in turn equips \modelflash{} with strong capabilities for controllable synthesis.

\subsection{Video Data}

We enhance the base video corpus from \model{}~\citep{ai2025ming} with two core data strategies to improve reasoning quality and interactive realism.

\textbf{High-Value Video Selection:}
We discard simple heuristics (\textit{e.g.}, resolution) and instead use a VLM with Chain-of-Thought reasoning to score videos across presentation, content, and reasoning difficulty. Clips strong in any dimension are retained, preserving challenging edge cases while filtering out bland, high-resolution filler.

\textbf{Realistic Multi-Turn Interaction Synthesis:} We generate temporally grounded dialogues for long videos (up to 30 min, 256k tokens) by combining hierarchical event extraction with a “look-back” mechanism that enforces cross-turn consistency. To model real user behavior, each dialogue is conditioned on an LLM-simulated cognitive state (\textit{e.g.}, curiosity or confusion), yielding intent-rich, multi-turn interactions beyond standard benchmarks.

%\textcolor{red}{Video streaming conversation constitutes a fundamental capability for MLLMs in video understanding. However, acquiring large-scale streaming multi-turn conversation data is prohibitively expensive. In this work, we propose a pipeline to systematically synthesize diverse, balanced, and high-quality multi-turn conversation video datasets.
%We collect 8.2M videos from the internet, ranging from 90 seconds to 10 minutes, and first filter out low-quality videos with high speech density, high shot density, irregular aspect ratios, or low resolution. We then filter out low-information, incoherent, or overly simple videos using SOTA MLLMs. %Qwen2.5-VL~\citep{bai2025qwen25vltechnicalreport}.
%To ensure a balanced dataset, we use advanced embedding models (\emph{e.g.}, Qwen3-Embedding~\citep{zhang2025qwen3} and M3-Embedding~\citep{chen2024bge}) to extract embeddings and then cluster the videos to suppress high-frequency data while preserving long-tail content. Finally, we use SOTA video understanding models to generate high-quality video conversations. This produces 1.2M conversation turns across 5-minute average videos, balanced across various task categories. }

\subsection{Text Data}
For text data, we utilize corpus from Ling~\citep{ling}, M2-Omni~\citep{guo2025m2}, and \model{}~\citep{ai2025ming} to preserve and further enhance the model’s language proficiency.

\section{Evaluation}
In this section, we present the evaluation details and quantitative examples of \modelflash{} on both public and in-house benchmarks.

\subsection{Public Benchmarks}

The details of the public benchmarks are provided in Appendix~\ref{sec:app_open_benchmark}. As shown in Table \ref{table_1_image2text}$\sim$\ref{table_11_streamingmultiturnbench}, our holistic assessment covers more than 50 rigorously curated public benchmarks across the following seven distinct multi-modal dimensions: Image $\rightarrow$ Text (Understanding), Text $\rightarrow$ Image (Generation), Image $\rightarrow$ Image (Editing), Image $\rightarrow$ Image (Segmentation), Audio $\rightarrow$ Text (Understanding), Text $\rightarrow$ Audio (Generation), and Video $\rightarrow$ Text (Understanding).

% ============= subsection =============
\subsection{In-house Benchmarks}
\label{sec:B_5_2_inhouse}

In addition to public benchmarks, we also establish three in-house benchmarks to comprehensively evaluate multiple capabilities of MLLMs, including:

\noindent\textbf{Wiki Knowledge}. To thoroughly evaluate the factual knowledge capabilities of our models, we have developed an in-house benchmark called Wiki Knowledge, constructed from Wikipedia. This benchmark encompasses both broad general knowledge and specialized domain categories, including famous landmarks, notable figures, and distinctive dishes, enabling a comprehensive assessment of MLLMs’ understanding and reasoning over diverse factual information.

\noindent\textbf{Video Streaming Multi-turn Benchmark}.
The evaluation of video streaming multi-turn dialogue capabilities requires quantifying not only the model's understand capability but also assessing its interactive experience, including proactivity and naturalness. Previous streaming dialogue datasets, such as StreamBench~\citep{lin2024streamingbench} and OvO-Bench~\citep{niu2025ovo}, have primarily focused on the understanding aspect while lacking a thorough evaluation of the interactive experience.
To address this gap, we introduce StreamingMultiturnBench. 
To construct StreamingMultiturnBench, we manually selected 380 videos, carefully ensuring coverage of multiple key domains including life recording, education, TV shows, video games, and documentaries. Then we use SOTA closed-source model%Gemini2.5 Pro~\citep{comanici2025gemini25}  
for machine annotation. Subsequently, a team of 10 human annotators revise and double-check the dialogue content to ensure it aligns with human conversational preferences. This process yielded 2,200 video question-answer pairs.
During evaluation, we use advanced closed-source model, \emph{e.g.} GPT-4o~\citep{chatgpt4o}, to compare the model's output against the human-annotated answers, scoring it on a scale of 1 to 5 across the five dimensions: accuracy, completeness, relevance, naturalness, and proactivity. The final score is the average for each dimension. To align our metrics with other video benchmarks, we linearly scale the results to a 100-point scale. We commit to open-sourcing and publicly maintaining this benchmark to ensure reproducibility.

\noindent\textbf{Multi-Dialect and Multi-Domain Audio Understanding Benchmark}. To extend audio understanding benchmarks into multi-dialect and multi-domain settings, we constructed two specialized datasets. The multi-dialect dataset was created from 15 regions, while the multi-domain one was curated from six domains. All samples were manually verified for quality by trained annotators. The final datasets comprise 51,986 multi-dialect samples and 10,397 multi-domain samples, with the latter distributed across: Noisy (8,145), Chat (443), Government (462), Health (450), Knowledge (421), and Local Services (476).

% --- 自定义设置 ---

% 2. 定义一个命令，用于将最佳分数加粗并添加下划线
%    这样可以使表格代码更清晰，也便于将来修改样式
\newcommand{\best}[1]{\textbf{\underline{#1}}}

\begin{table*}[bthp]

\begin{minipage}[t]{1.0\linewidth}
        % ***--***--*** 1. OC Normal
        \caption{
            Performance of \modelflash{} on \textbf{Vision-to-Text Benchmarks} compared to leading models. * denotes our own evaluation using the official benchmark prompts. 
        }
\centering
\small

\setlength{\tabcolsep}{3.5pt}
\begin{tabular}{l l | c | c c  c | c}
\hline
% --- 表头 ---
Type & Benchmark & \makecell{Ming-Flash \\ Omni} & \makecell{Qwen3-Omni \\ 30B-A3B} & \makecell{LongCat-Flash-Omni \\ 560B-A27B} & \makecell{Gemini2.5 \\ Pro} & \makecell{GLM-4.6-V \\ 106B-A12B} \\
\hline
% --- General 部分 ---
\multirow{5}{*}{General} 
& MMStar & 74.9 & 68.5 & 70.9 & 73.6 & \best{75.9*} \\ 
& AI2D & 89.3 & 85.2 & - & \best{89.5} & 88.8* \\
& HallusionBench & \best{66.1} & 59.7 & - & 64.1 & 63.2* \\
& MMVet & \best{85.6} & 73.9* & 69.0 & 83.3* & 79.8* \\ 
& $\text{MMBench-EN}_{\text{test}}$ & 87.0 & 84.7* & 87.5 & \best{90.8}* & 88.8* \\ \hline
\multirow{1}{*}{Knowledge} 
& Wiki Knowledge(In-house) & \best{64.6} & 36.5 & - & 48.7 & 35.3 \\ \hline
% --- STEM/Reasoning 部分 ---
\multirow{5}{*}{STEM/Reasoning} 
& MMMU\textsubscript{val} & 77.1 & 69.1 & 70.7 & \best{80.9} & 76.0 \\
%& MMMU-Pro & 61.4 & 57.0 & - & \best{71.2} & 66.0 \\
& MathVista\textsubscript{mini} & 83.9 & 75.9 & 77.9 & 77.7 & \best{85.2} \\
& MathVerse\textsubscript{Vision} & \best{75.5} & 49.6 & - & 73.1 & 75.4 \\
& MathVision & 65.6 & 56.3 & - & \best{66.0} & 63.5 \\
& LogicVista & 67.6 & 48.3 & - & \best{68.7} & 62.1 \\ \hline
% --- OCR 部分 ---
\multirow{3}{*}{OCR} 
& ChartQA & 87.0 & 87.5 & 87.6 & 83.3 & \best{91.1} \\
& DocVQA & 94.7 & 95.0 & 91.8 & - & \best{95.6} \\
& OCRBench & \best{891} & 860 & 849 & 866 & 865\\ \hline
% --- Grounding 部分 ---
\multirow{1}{*}{Grounding}
& RefCOCO-avg & \best{88.5} & 87.4 & - & 87.5 & 85.9 \\ \hline
% --- Multi-image 部分 ---
\multirow{3}{*}{Multi-image} 
& MMTBench\_val\_mi & 70.1 & 69.9* & - & 69.7* & \best{70.6}* \\
& MuirBench & 61.4 & 61.9 & - & - & \best{77.1} \\
& LLaVA-Interleave & 63.1 & 61.6* & - & - & \best{63.5}* \\ \hline
% --- Video 部分 ---
\multirow{7}{*}{Video}
& MVBench & \best{77.5} & 71.2* & 75.2 & 74.2* & 74.9 \\
& VideoMME \textit{w/o\ sub} & 73.4 & 70.5 & 76.2 & \best{84.3} & 74.8 \\
& LongVideoBench & 65.4 & 59.4* & \best{69.3} & 67.1* & 66.5* \\ 
& MLVU & 77.7 & 75.2 & - & \best{81.2}* & 75.1* \\
& PerceptionTest & 78.5 & 65.5* & - & 78.4 & \best{80.6}* \\
& Charades\_STA & \best{59.3} & 28.9* & - & - & 27.1* \\
& TOMATO & 40.4 & 28.2* & - & \best{46.9}* & 39.4* \\ \hline
\label{table_1_image2text}
\end{tabular}

\end{minipage}
\end{table*}

\begin{table*}

\begin{minipage}[t]{1.0\linewidth}
        % ***--***--*** 4. Image Generation
        \caption{
            Performance of \modelflash{} on \textbf{Text-to-Image Generation Benchmarks} compared to leading models. ``\textit{Gen.}'' denotes models for pure image generation, while ``\textit{Uni.}'' denotes models capable of both image understanding and generation. Note that the global best performance is highlighted by an \underline{underline}, and the local best result in ``\textit{Gen.}'' or ``\textit{Uni.}'' is marked with \textbf{bold}.
        }
	\small
	\vspace{-0.2cm}
	\centering
        \begin{tabular}{
            p{0.8cm}<{\raggedright}p{3.2cm}<{\raggedright}|p{0.8cm}<{\centering}p{0.8cm}<{\centering}p{0.8cm}<{\centering}p{0.8cm}<{\centering}p{0.8cm}<{\centering}p{0.8cm}<{\centering}p{0.8cm}<{\centering}|p{1.4cm}<{\centering}
        }
            \hline
            \multicolumn{1}{l}{\multirow{2}{*}{Type}} &
            \multicolumn{1}{l|}{\multirow{2}{*}{Model}} &\multicolumn{7}{c|}{GenEval} &
             \multicolumn{1}{c}{\multirow{2}{*}{DPG-Bench}} \\ 
             \cline{3-9}
            &&\multicolumn{1}{c}{1-Obj.} &\multicolumn{1}{c}{2-Obj.} &
            \multicolumn{1}{c}{Count} &
            \multicolumn{1}{c}{Colors} &
            \multicolumn{1}{c}{Posit.} &
            \multicolumn{1}{c}{Color.} &
            \multicolumn{1}{c|}{AVG} &\\
            \hline
            % \multicolumn{7}{c}{\multirow{1}{*}{ \textit{Methods: Video-text Pre-training from Scratch}}} \\\hline
            % & LlamaGen & 0.71 & 0.34 & 0.21 & 0.58 & 0.07 & 0.04 & 0.32 & - \\
            % & LDM & 0.92 & 0.29 & 0.23 & 0.70 & 0.02 & 0.05 & 0.37 & - \\
            % & SDv1.5 &  0.97 & 0.38 & 0.35 & 0.76 & 0.04 & 0.06 & 0.43 & - \\
            % & PixArt-$\alpha$ &  0.98 & 0.50 & 0.44 & 0.80 & 0.08 & 0.07 & 0.48 & - \\
            % \multicolumn{1}{l}{\textit{Gen.}}
            \multirow{5}{*}{\textit{Gen.}} & SDv2.1 &  0.98 & 0.51 & 0.44 & 0.85 & 0.07 & 0.17 & 0.50 & 68.09 \\
            & Emu3-Gen & 0.98 & 0.71 & 0.34 & 0.81 & 0.17 & 0.21 & 0.54 & 80.60 \\
            & SDXL & 0.98 & 0.74 & 0.39 & 0.85 & 0.15 & 0.23 & 0.55 & 74.65 \\
            & DALL-E 3 & 0.96 & 0.87 & 0.47 & 0.83 & \textbf{0.43} & 0.45 & 0.67 & - \\
            & SD3-Medium  & \textbf{0.99} & \textbf{0.94} & \textbf{0.72} & \textbf{0.89} & 0.33 & \textbf{0.60} & \textbf{0.74} & \textbf{84.08} \\
            \hline
            % \multicolumn{1}{l}{\textit{Uni.}}
            \multirow{13}{*}{\textit{Uni.}}
            % & LWM &  0.93 & 0.41 & 0.46 & 0.79 & 0.09 & 0.15 & 0.47 & - \\
            & SEED-X & 0.97 & 0.58 & 0.26 & 0.80 & 0.19 & 0.14 & 0.49 & - \\
            & Show-o &  0.95 & 0.52 & 0.49 & 0.82 & 0.11 & 0.28 & 0.53 & - \\
            & TokenFlow-XL &  0.95 & 0.60 & 0.41 & 0.81 & 0.16 & 0.24 & 0.55 & - \\
            & Janus & 0.97 & 0.68 & 0.30 & 0.84 & 0.46 & 0.42 & 0.61 & 79.68 \\
            & JanusFlow &  0.97 & 0.59 & 0.45 & 0.83 & 0.53 & 0.42 & 0.63 & 80.09 \\
            & JanusPro-7B &  0.99 & 0.89 & 0.59 & 0.90 & 0.79 & 0.66 & 0.80 & 84.19 \\
            & UniWorld-V1 &  0.98 & 0.93 & 0.81 & 0.89 & 0.74 & 0.71 & 0.84 & 81.38 \\
            & OmniGen2 & 0.99 & 0.96 & 0.74 & \underline{\textbf{0.98}}  & 0.71 & 0.75 & 0.86 & 83.57 \\
            & BAGEL &  0.99 & 0.94 & 0.81 & 0.88 & 0.64 & 0.63 & 0.82 & - \\
            & Z-Image &  \underline{\textbf{1.00}} & 0.94 & 0.78 & 0.93 & 0.62 & 0.77 & 0.84 & 88.14 \\
            & Qwen-Image &  0.99 & 0.92 & 0.89 & 0.88 & 0.76 & 0.77 & 0.87 & \underline{\textbf{88.32}} \\
            % \rowcolor{gray!15}
            % \cellcolor{white} 
            & Qwen-Image-RL &  \underline{\textbf{1.00}} & 0.95 & \underline{\textbf{0.93}} & 0.92 & 0.87 & 0.83 & 0.91 & - \\
            \cdashline{2-10} % 
            & \modelflash{} & 0.99 & \underline{\textbf{0.98}} & 0.92 & 0.94 & \underline{\textbf{0.96}} & \underline{\textbf{0.89}} & \underline{\textbf{0.94}} & 86.98 \\
            \hline
            
            \label{table_g_imggen}
        \end{tabular}

        \vspace{0.4cm}
\end{minipage}

\end{table*}

\begin{table*}[thbp]
\vspace{-0.1cm}
\centering
\small
\caption{
    % Performance of \modelflash{} on \textbf{Image-to-Image Editing Benchmarks} compared to leading models. All metrics are evaluated by GPT-4.1. ``\textit{Edit.}'' denotes models specifically trained for image editing, while ``\textit{Unified.}'' denotes models capable of image understanding, generation and editing. The global best performance is marked with \textbf{bold}.
    Performance of \modelflash{} on \textbf{Image-to-Image Editing Benchmarks} compared to leading models. All metrics are evaluated by GPT-4.1. ``\textit{Edit.}'' denotes models specifically trained for image editing, while ``\textit{Generalist.}'' denotes models that serve as general-purpose models capable of image understanding, generation, and editing. The global best performance is marked with \textbf{bold}.
    % Best results are highlighted in \textbf{bold}.
    % 最好的结果加粗表示，次优的使用下划线
    % ``\textit{Gen.}'' denotes models for pure image generation, while ``\textit{Uni.}'' denotes models capable of both image understanding and generation.
}
\begin{tabular}{@{}ll|ccc|ccc@{}}
\toprule
\multirow{2}{*}{Type}              & \multirow{2}{*}{Model} & \multicolumn{3}{c|}{GEdit-Bench-EN (Full set)$\uparrow$} & \multicolumn{3}{c}{GEdit-Bench-CN (Full set)$\uparrow$} \\
                                   &                        & G\_SC          & G\_PQ         & G\_O          & G\_SC         & G\_PQ         & G\_O          \\ \midrule
\multirow{6}{*}{\textit{Edit.}}    & Instruct-Pix2Pix       & 3.58           & 5.49          & 3.68          & -             & -             & -             \\
                                   & AnyEdit                & 3.18           & 5.82          & 3.21          & -             & -             & -             \\
                                   & MagicBrush             & 4.68           & 5.66          & 4.52          & -             & -             & -             \\
\multicolumn{1}{l}{}               & Step1X-Edit            & 7.09           & 6.76          & 6.70          & 7.20          & 6.87          & 6.86          \\
\multicolumn{1}{l}{}               & Qwen-Image-Edit        & 8.00  & 7.86 & 7.56 & 7.82 & 7.79 & 7.52 \\ 
\multicolumn{1}{l}{}               & Z-Image-Edit        & \textbf{8.11}  & 7.72 & 7.57 & \textbf{8.03} & 7.80 & 7.54 \\ 
\midrule
\multirow{5}{*}{\textit{Generalist.}} & UniWorld-v1            & 4.93           & 7.43 & 4.85          & -             & -             & -             \\
                                   & OmniGen                & 5.96           & 5.89          & 5.06          & -             & -             & -             \\
                                   & OmniGen2               & 7.16           & 6.77          & 6.41          & -             & -             & -             \\
                                   & BAGEL       & 7.36           & 6.83          & 6.52          & 7.34 & 6.85          & 6.50          \\
                                   & \modelflash{}          & \textbf{8.11} & \textbf{7.87}          & \textbf{7.64} & 8.02         & \textbf{7.95} & \textbf{7.62} \\ \bottomrule
\end{tabular}
\label{table_i2i_gedit}
% \end{table*}
\vspace{0.4cm}
% \begin{table*}[htbp]
\centering
\small
\caption{
    Performance of \modelflash{} on \textbf{Image-to-Mask Segmentation Benchmarks} compared to leading models. 
    Model types are denoted as: \textit{Vision.} for vision-only models, \textit{SAM.} for models equipped with an additional SAM-like segmentation head, and \textit{Uni.} for unified MLLMs capable of both understanding and generation. Results with ``*'' are obtained by evaluating on 500 images sampled from each dataset via the official API.
    % \textit{Vision.}表示纯视觉模型，\textit{SAM.}表示使用了 类似 SAM 的额外分割 head，\textit{Unified.}表示统一理解与生成MLLM。* 表示我们使用了官方 api 在每个数据集里采样了 500 张图像进行评估
    % ``\textit{Gen.}'' denotes models for pure image generation, while ``\textit{Uni.}'' denotes models capable of both image understanding and generation. Note that the global best performance is highlighted by an \underline{underline}, and the local best result in ``\textit{Gen.}'' or ``\textit{Uni.}'' is marked with \textbf{bold}.
}
\begin{tabular}{@{}cl|ccc@{}}
\toprule
Type                              & Model        & RefCOCO (val)$\uparrow$ & RefCOCO+ (val)$\uparrow$ & RefCOCOg (val)$\uparrow$ \\ \midrule
\multirow{4}{*}{\textit{Vision.}} & VLT          & 67.5          & 56.3           & 55.0           \\
                                  & CRIS         & 70.5          & 62.3           & 59.9           \\
                                  & LAVT         & 72.7          & 62.1           & 61.2           \\
                                  & PolyFormer-B & 74.8          & 67.6           & 67.8           \\ \midrule
\multirow{3}{*}{\textit{SAM.}}    & LISA-7B      & 74.1          & 62.4           & 66.4           \\
                                  & PixelLM-7B   & 73.0          & 66.3           & 69.3           \\
                                  & OMG-LLAVA    & 75.6          & 65.6           & 70.7           \\ \midrule
\multirow{3}{*}{\textit{Uni.}}     & Nano-banana*     & 15.7                    & 13.9           & 14.9           \\
                                   & Qwen-Image-Edit* & 30.3                    & 28.8           & 34.0           \\
                                   & \modelflash{}       & 72.1                    & 65.2           & 65.4           \\ \bottomrule
\end{tabular}
\label{table_i2i_seg}
% \end{table*}
\vspace{0.4cm}

        \captionsetup{font={small}}
        \caption{
            Performance of \modelflash{} on \textbf{PUBLIC Text-to-Speech Benchmarks} compared to leading MLLMs.
        }
        \small
	\vspace{-0.2cm}
	\centering
        \begin{tabular}{
p{0.4cm}<{\centering}p{2.2cm}<{\centering}|p{1.2cm}<{\centering}p{1.2cm}<{\centering}p{1.2cm}<{\centering}p{1.2cm}<{\centering}p{1.2cm}<{\centering}p{1.2cm}<{\centering}p{1.2cm}<{\centering}p{1.2cm}<{\centering}
}
\hline
\multicolumn{1}{c}{\multirow{2}{*}{Type}} &
\multicolumn{1}{c|}{Benchmark} &
\multicolumn{1}{c}{Ming-Flash} &
\multicolumn{1}{c}{Ming-Lite} &
\multicolumn{1}{c}{Qwen3} &
\multicolumn{1}{c}{Seed} &
\multicolumn{1}{c}{F5} &
\multicolumn{1}{c}{CosyVoice3} &
\multicolumn{1}{c}{Qwen2.5}
\\
&
\multicolumn{1}{c|}{(Seed-TTS-Eval)} &
\multicolumn{1}{c}{Omni} &
\multicolumn{1}{c}{Omni} &
\multicolumn{1}{c}{Omni} &
\multicolumn{1}{c}{TTS} &
\multicolumn{1}{c}{TTS} &
\multicolumn{1}{c}{} &
\multicolumn{1}{c}{Omni}
\\ \hline
% \multicolumn{7}{c}{\multirow{1}{*}{ \textit{Methods: Video-text Pre-training from Scratch}}} \\\hline
\multicolumn{1}{c}{\multirow{2}{*}{\textit{Chinese}}} & Zh-wer $\downarrow$ & \textbf{0.87} & 1.69 & 1.07 & 1.11 & 1.56 & 1.16 & 1.70
\\ % \hline
& Zh-sim $\uparrow$ & 0.72 & 0.68 & - & \textbf{0.80} & 0.74 & 0.78 & 0.75
\\ \hline
\multicolumn{1}{c}{\multirow{2}{*}{\textit{English}}} & En-wer $\downarrow$ & 2.19 & 4.31 & \textbf{1.39} & 2.24 & 1.83 & 2.02 & 2.72
\\ % \hline
& En-sim $\uparrow$ & 0.61 & 0.51 & - & \textbf{0.76} & 0.65 & 0.718 & 0.63
\\ \hline

\label{table_10_audiotts}
\end{tabular}

% \end{table*}

% ---- sichuan dialect result ----- %
\centering
\captionsetup{justification=centering}
\caption{Performance of \modelflash{} on Voice Control for Sichuanese Dialect Generation benchmarks.}
\label{tab::audiotts_dialect_sichuanese}
% \resizebox{\textwidth}{!}{%
\begin{tabular}{lcc}
\toprule
\textbf{Model} & \multicolumn{2}{c}{\textbf{Performance}} \\
\midrule
& \multicolumn{1}{c}{\textbf{WSC-TTS-Eval-easy}} &
\multicolumn{1}{c}{\textbf{WSC-TTS-Eval-hard}} \\
& \multicolumn{1}{c}{\footnotesize \textbf{CER(\%)$\downarrow$ | SIM $\uparrow$ | ACC(\%) $\uparrow$ }} &
\multicolumn{1}{c}{\footnotesize \textbf{CER(\%)$\downarrow$ | SIM $\uparrow$ | ACC(\%) $\uparrow$ }} \\

\cmidrule(lr){2-2} \cmidrule(lr){3-3}
Cosyvoice3      & 3.17 | 0.696 | 68.06 & 4.07 | \textbf{0.723} | 80.90 \\
Step-Audio-TTS      & 10.83 | 0.676 | --- & 12.52 | 0.545 | --- \\
Qwen-TTS      & 4.13 | --- | --- & 7.35 | --- | ---  \\
CosyVoice2-WSC  & 4.28 | 0.727 | --- & 8.78 | 0.625 | --- \\
CosyVoice2-WSC-SFT  &  4.08 | \textbf{0.788} | --- & 7.22 | 0.679 | --- \\
Ming-Flash-Omni         &  \textbf{2.25} | 0.695 | \textbf{82.08}   & \textbf{3.18} | 0.717 | \textbf{84.42}   \\

\bottomrule
\end{tabular}%
% } % resizebox

\vspace{-0.1cm}
% \begin{table*}[!t]

\end{table*}

% ----  result of podcast -----%
\begin{table}[h!]
\centering
\small
\caption{Performance of Ming-Flash-Omni on on ZipVoice-Dia(zh) test sets. Arrows indicate the desired direction ($\downarrow$ = lower is better, $\uparrow$ = higher is better). Best values per column are in \textbf{bold}. }
\label{tab:podcast_results}
\begin{tabular}{l ccc}
\toprule
\multirow{2}{*}{\textbf{Model}} & \multicolumn{3}{c}{\textbf{ZipVoice-Dia(zh)}} \\
\cmidrule(lr){2-4}
 & \textbf{CER-zh $\downarrow$} & \textbf{cpSIM $\uparrow$} & \textbf{UTMOS $\uparrow$} \\
\midrule
ZipVoice-Dia  & 3.39 & 0.553 & 2.24 \\
MoonCast~\citep{ju2025mooncast}  & 27.43 & 0.441 & 1.76 \\
MOSS-TTSD~\citep{zhao2025moss}  & 8.62 & 0.421 & 1.70 \\
VibeVoice-1.5B~\citep{peng2025vibevoice}  & 12.87 & 0.455 & 1.74 \\
FireRedTTS2~\citep{xie2025fireredtts}  & 3.34 & 0.512 & 1.90 \\
SoulX-Podcast~\citep{xie2025soulx}  & 2.2 & \textbf{0.599} & 2.09 \\
Ming-Flash-Omni & \textbf{2.12} & 0.457 & \textbf{2.25} \\
\bottomrule
\end{tabular}
\end{table}

% in preamble:
% \usepackage{booktabs}
% \usepackage{graphicx}
% \usepackage{caption}

\begin{table*}[!t]
\centering
\captionsetup{justification=centering}
\caption{Performance of \modelflash{} on Context ASR benchmarks.}
\label{tab:main_context_asr_perf}

\resizebox{\textwidth}{!}{%
\begin{tabular}{@{}llccccc@{}}
\toprule
& \textbf{Model} & \multicolumn{5}{c}{\textbf{Performance}} \\
\midrule
& &
\multicolumn{1}{c}{\textbf{Avg}} &
\multicolumn{1}{c}{\textbf{Speech-English}} &
\multicolumn{1}{c}{\textbf{Dialogue-English}} &
\multicolumn{1}{c}{\textbf{Speech-Mandarin}} &
\multicolumn{1}{c}{\textbf{Dialogue-Mandarin}} \\
& &
\multicolumn{1}{c}{} &
\multicolumn{1}{c}{\footnotesize \textbf{WER | NE-WER | NE-FNR}} &
\multicolumn{1}{c}{\footnotesize \textbf{WER | NE-WER | NE-FNR}} &
\multicolumn{1}{c}{\footnotesize \textbf{WER | NE-WER | NE-FNR}} &
\multicolumn{1}{c}{\footnotesize \textbf{WER | NE-WER | NE-FNR}} \\
\cmidrule(lr){3-3} \cmidrule(lr){4-4} \cmidrule(lr){5-5} \cmidrule(lr){6-6} \cmidrule(lr){7-7}
& Gemini2.5-Pro       & 6.3& 6.65 | 12.82 | 8.86& 3.59 | 8.18 | 3.53& 5.37 | 8.35 | 4.09& 4.03 | 8.96 | 1.18\\
& Kimi-Audio          & 9.24&  2.90 |  6.68 |  8.01 &  4.67 | 13.50 | 11.31 & 1.95 | 11.13 | 15.28 & 2.90 | 15.91 | 16.68 \\
& Baichuan-Omni-1.5   & 8.17&  8.16 |  7.69 |  6.53 &  9.91 | 14.40 |  5.54 & 2.98 |  8.39 |  4.71 & 5.00 | 16.83 |  7.84 \\
& Qwen3-ASR& 4.23&  2.95 | 5.00 | 3.93&  \textbf{2.91} | 8.19 | 4.38& 0.92 | 5.92 | 1.17& \textbf{1.36} | 10.18 | 3.83\\
& Qwen3-Omni-30B-A3B-Instruct& 4.34&  \textbf{1.13} | \textbf{1.66} | \textbf{0.28}&  4.41 | 21.26 | \textbf{1.15}& \textbf{0.80} | \textbf{5.41} | \textbf{0.36}& 1.59 | 13.35 | \textbf{0.69}\\
& Ming-Flash-Omni     & \textbf{3.30}& 2.91 | 2.53 | 2.13& 3.50 | \textbf{7.60} | 2.08& 1.13 | \textbf{5.41} | 1.16& 1.77 | \textbf{8.30} | 1.04\\
\bottomrule
\end{tabular}%
} % resizebox
\end{table*}

\begin{table*}[bthp]

        % ***--***--*** 8-1. Audio ASR Open Source
        \captionsetup{font={small}}
        \caption{
            Performance of \modelflash{} on \textbf{PUBLIC and IN-HOUSE Audio Understanding Benchmarks}.
        }
        \small
	\vspace{-0.2cm}
	\centering
        \begin{tabular}{
            p{1.6cm}<{\centering}p{8.0cm}<{\centering}p{1.6cm}<{\centering}p{1.6cm}<{\centering}p{1.6cm}<{\centering}p{1.6cm}<{\centering}<{\centering}p{1.6cm}<{\centering}
        }
            \hline
            \multicolumn{1}{c}{\multirow{2}{*}{Type}} &
		\multicolumn{1}{c|}{\multirow{2}{*}{Benchmark}} &
            \multicolumn{1}{c}{Ming-Flash} &
            \multicolumn{1}{c}{Qwen3} &
        \multicolumn{1}{c}{Qwen2} &
            \multicolumn{1}{c}{Kimi}
		\\
            &
            \multicolumn{1}{c|}{} &
		\multicolumn{1}{c}{Omni} &
        \multicolumn{1}{c}{Omni} &
		\multicolumn{1}{c}{Audio} &
		\multicolumn{1}{c}{Audio}
		\\ \hline
        % \multicolumn{7}{c}{\multirow{1}{*}{ \textit{Methods: Video-text Pre-training from Scratch}}} \\\hline
        & \multicolumn{1}{c|}{Aishell1 $\downarrow$} & 1.11 & 1.04 & 1.53 & \underline{\textbf{0.60}}
		\\ % \hline
        & \multicolumn{1}{c|}{Aishell2-test-android $\downarrow$} & \underline{\textbf{2.41}}  & 2.64  & 2.92 & 2.64
		\\ % \hline
        \textit{PUBLIC} & \multicolumn{1}{c|}{Aishell2-test-ios $\downarrow$} & \underline{\textbf{2.44}}  & 2.55 & 2.92 & 2.56
		\\ % \hline
        \textit{Chinese} & \multicolumn{1}{c|}{Cv15-zh $\downarrow$} & 5.03 & \underline{\textbf{4.31}} & 6.90 & 7.21
		\\ % \hline
        \textit{Benchmarks} & \multicolumn{1}{c|}{Fleurs-zh $\downarrow$} & 2.82 & \underline{\textbf{2.20}}  & 7.50 & 2.69
		\\ % \hline
        & \multicolumn{1}{c|}{Wenetspeech-testmeeting $\downarrow$} & \underline{\textbf{5.83}} & 5.89 & 7.16 & 6.28
            \\ % \hline
        & \multicolumn{1}{c|}{Wenetspeech-testnet $\downarrow$} & 5.02 & \underline{\textbf{4.69}}  & 8.42 & 5.37
        \\ % \hline
        & \multicolumn{1}{c|}{SpeechIO $\downarrow$} & 2.49 & \underline{\textbf{2.18}}  & 3.01 & 2.23
            \\ 
            \cdashline{2-7}
        & \multicolumn{1}{c|}{Average (Chinese) $\downarrow$} & 3.39  & \underline{\textbf{3.19}}  & 5.05 & 3.70
            \\ 
            \hline
        & \multicolumn{1}{c|}{Librispeech-test-clean $\downarrow$} & \underline{\textbf{1.17}} & 1.22  & 1.60 & 1.28
		\\ % \hline
        & \multicolumn{1}{c|}{Librispeech-test-other $\downarrow$} & \underline{\textbf{2.20}} & 2.48  & 3.60 & 2.42
		\\ % \hline
        \textit{PUBLIC} & \multicolumn{1}{c|}{Multilingual-librispeech $\downarrow$} & 3.75 & \underline{\textbf{3.67}}  & 5.40 & 5.88
		\\ % \hline
        \textit{English} & \multicolumn{1}{c|}{Cv15-en $\downarrow$} & \underline{\textbf{5.99}} & 6.05  & 8.60 & 10.31
		\\ % \hline
        \textit{Benchmarks} & \multicolumn{1}{c|}{Fleurs-en $\downarrow$} & 2.99 & \underline{\textbf{2.72}}  & 6.90 & 4.44
		\\ % \hline
        & \multicolumn{1}{c|}{Voxpopuli-v1.0-en $\downarrow$} & \underline{\textbf{5.76}} & 6.02  & 6.84 & 7.97
		\\ 
            \cdashline{2-7}
        & \multicolumn{1}{c|}{Average (English) $\downarrow$} & \underline{\textbf{3.64}}  & 3.69  & 5.49 & 5.38
            \\ 
        \hline
        & \multicolumn{1}{c|}{Hunan $\downarrow$}  & \underline{\textbf{7.63}} & 20.82  & 25.88 & 31.93
		\\ % \hline
        & \multicolumn{1}{c|}{Minnan $\downarrow$}  & \underline{\textbf{13.29}} & 24.60  & 123.78 & 80.28
		\\ % \hline
        & \multicolumn{1}{c|}{Guangyue $\downarrow$}  & \underline{\textbf{4.21}} & 27.43  & 7.59 & 41.49
		\\ % \hline
        & \multicolumn{1}{c|}{Chuanyu $\downarrow$}  & \underline{\textbf{3.80}} & 5.54  & 7.77 & 6.69
		\\ % \hline
        & \multicolumn{1}{c|}{Shanghai $\downarrow$}  & \underline{\textbf{10.19}} & 30.96  & 31.73 & 60.64
        \\ % \hline
        & \multicolumn{1}{c|}{Anhui $\downarrow$}  & \underline{\textbf{5.49}} & 5.70  & 5.72 & 8.61
        \\ % \hline
        \textit{IN-HOUSE} & \multicolumn{1}{c|}{Dongbei $\downarrow$}  & 5.14 & \underline{\textbf{3.92}}  & 4.87 & 4.40
        \\ % \hline
        \textit{Dialect} & \multicolumn{1}{c|}{Henan $\downarrow$}  & \underline{\textbf{7.47}} & 8.94  & 12.31 & 14.40
        \\ % \hline
        \textit{Benchmarks} & \multicolumn{1}{c|}{Hubei $\downarrow$}  & \underline{\textbf{6.14}} & 14.55  & 16.39 & 20.07
        \\ % \hline
        & \multicolumn{1}{c|}{Jiangsu $\downarrow$}  & \underline{\textbf{10.46}} & 13.19  & 12.80 & 17.25
        \\ % \hline
        & \multicolumn{1}{c|}{Kejiahua $\downarrow$}  & \underline{\textbf{16.87}} & 27.88  & 22.33 & 29.78
        \\ % \hline
        & \multicolumn{1}{c|}{Shaanxi $\downarrow$}  & 6.48 & \underline{\textbf{5.31}}  & 6.32 & 6.09
        \\ % \hline
        & \multicolumn{1}{c|}{Shandong $\downarrow$}  & \underline{\textbf{10.32}} & 17.04  & 15.00 & 18.66
        \\ % \hline
        & \multicolumn{1}{c|}{Tianjin $\downarrow$}  & \underline{\textbf{13.33}} & 19.70  & 21.78 & 34.57
        \\ % \hline
        & \multicolumn{1}{c|}{Yunnan $\downarrow$}  & \underline{\textbf{10.44}} & 19.78  & 21.57 & 32.79
		\\ \cdashline{1-7}
        & \multicolumn{1}{c|}{Noisy $\downarrow$}  & 11.15 & \underline{\textbf{9.25}}  & 12.46 & 24.40
        \\ % \hline
        \textit{IN-HOUSE} & \multicolumn{1}{c|}{Chat $\downarrow$}  & 2.96 & \underline{\textbf{2.09}}  & 4.29 & 2.96
		\\ % \hline
        \textit{Domain} & \multicolumn{1}{c|}{Government $\downarrow$}  & 1.57  & \underline{\textbf{1.55}}  & 2.70 & 2.03
		\\ % \hline
        \textit{Benchmarks} & \multicolumn{1}{c|}{Health $\downarrow$}  & 3.15 & \underline{\textbf{2.22}}  & 4.18 & 2.38
		\\ % \hline
        & \multicolumn{1}{c|}{Knowledge $\downarrow$}  & 3.17 & 11.52   & 3.33 & \underline{\textbf{1.98}}
		\\ % \hline
        & \multicolumn{1}{c|}{Local-live $\downarrow$}  & 2.07 & \underline{\textbf{1.37}}  & 2.34 & 2.05
		\\ 
            \cdashline{1-7}
        \multicolumn{2}{c|}{Average (All IN-HOUSE) $\downarrow$} & \underline{\textbf{7.40}} & 13.02  & 17.39 & 21.12 
		\\ \hline

        \label{table_8_audioasr}
	\end{tabular}    
% \end{table*}

% \begin{table*}[bthp]
        % ***--***--*** 9. Audio QA
        \captionsetup{font={small}}
        \caption{
            Performance of \modelflash{} on \textbf{PUBLIC Audio Question-Answering Benchmarks}. }
        \small
	\vspace{-0.2cm}
	\centering
    \resizebox{\textwidth}{!}{%
        \begin{tabular}{
            p{1.8cm}<{\centering}p{0.9cm}<{\centering}p{1.0cm}<{\centering}p{1.2cm}<{\centering}|p{1.2cm}<{\centering}|p{1.0cm}<{\centering}p{1.2cm}<{\centering}|p{1.2cm}<{\centering}|p{1.2cm}<{\centering}
        }
            \hline
		\multicolumn{1}{c}{\multirow{2}{*}{Models}} &
        \multicolumn{1}{c|}{\multirow{2}{*}{Mean}} &
            \multicolumn{2}{c|}{Open-ended QA} &
            \multicolumn{1}{c|}{Knowledge} &
            \multicolumn{2}{c|}{Multi-Choice QA} &
            \multicolumn{1}{c|}{Instruction} &
            \multicolumn{1}{c}{Safety}
		\\ 
            & \multicolumn{1}{c|}{} &
		\multicolumn{1}{c}{AlpacaEval} &
            \multicolumn{1}{c|}{CommonEval} &
            \multicolumn{1}{c|}{SD-QA} &
            \multicolumn{1}{c}{MMSU} &
            \multicolumn{1}{c|}{OpenBookQA} &
            \multicolumn{1}{c|}{IFEval} &
            \multicolumn{1}{c}{AdvBench}
		\\ \hline
        % \multicolumn{7}{c}{\multirow{1}{*}{ \textit{Methods: Video-text Pre-training from Scratch}}} \\\hline
        \multicolumn{1}{c}{Step-Audio-chat} & \multicolumn{1}{c|}{57.10} & 79.80 & 59.80 & 46.84 & 31.87 & 29.19 & {65.77} & 86.73 \\
        \multicolumn{1}{c}{Qwen2-Audio-chat} & \multicolumn{1}{c|}{54.70} & 73.80 & 68.00 & 35.35 & 35.43 & 49.01 & 22.57 & 98.85 \\
        \multicolumn{1}{c}{Baichuan-Audio} & \multicolumn{1}{c|}{62.50} & 80.00 & 67.80 & 49.64 & 48.80 & 63.30 & 41.32 & 86.73 \\
        \multicolumn{1}{c}{GLM-4-Voice} & \multicolumn{1}{c|}{57.20} & 81.20 & 69.60 & 43.31 & 40.11 & 52.97 & 24.91 & 88.08 \\
        \multicolumn{1}{c}{Kimi-Audio} & \multicolumn{1}{c|}{76.90} & 89.20 & 79.40 & 63.12 & 62.17 & 83.52 & 61.10 & \underline{\textbf{100.00}} \\
        \cdashline{1-9}
        \multicolumn{1}{c}{Megrez-3B-Omni} & \multicolumn{1}{c|}{46.20} & 70.00 & 59.00 & 25.95 & 27.03 & 28.35 & 25.71 & 87.69 \\
        \multicolumn{1}{c}{DiVA} & \multicolumn{1}{c|}{55.70} & 73.40 & 70.80 & 57.05 & 25.76 & 25.49 & 39.15 & 98.27 \\
        \multicolumn{1}{c}{Qwen2.5-Omni} & \multicolumn{1}{c|}{74.10} & 89.80 & 78.60 & 55.71 & 61.32 & 81.10 & 52.87 & 99.42 \\
        \multicolumn{1}{c}{Qwen3-Omni-Flash-Instruct} & \multicolumn{1}{c|}{\underline{\textbf{85.40}}} & 95.40 & \underline{\textbf{91.00}} & \underline{\textbf{76.80}} & 68.40 & \underline{\textbf{91.40}} & \underline{\textbf{75.20}} & 99.40 \\
        \multicolumn{1}{c}{Baichuan-Omni-1.5} & \multicolumn{1}{c|}{71.10} & 90.00 & 81.00 & 43.40 & 57.25 & 74.51 & 54.54 & 97.31 \\
        \multicolumn{1}{c}{MiniCPM-o} & \multicolumn{1}{c|}{71.70} & 88.40 & 83.00 & 50.72 & 54.78 & 78.02 & 49.25 & 97.69 \\
        \cdashline{1-9}
        \multicolumn{1}{c}{\modelflash{}} & \multicolumn{1}{c|}{82.95} & \underline{\textbf{95.58}} & 87.10 & 71.30 & \underline{\textbf{72.09}} & 85.50 & 69.87 & 99.23 \\
        \hline

        \label{table_9_audioqa}
	\end{tabular}
    } % end \resizebox
\end{table*}

\begin{table*}[bthp]
        \centering
        \vspace{0.0cm}

        % ***--***--*** Streaming Multiturn Bench
    
        \captionsetup{font={small}}
        \caption{
            Performance of \modelflash{} on \textbf{StreamingMultiturnBench} compared to leading omni-MLLMs. 
        }
        \setlength{\tabcolsep}{8 pt}
        \label{table_11_streamingmultiturnbench}
        \small
	\vspace{-0.2cm}
	\centering
        \begin{tabular}{
            p{3.3cm}<{\centering}|p{1.1cm}<{\centering}p{1.4cm}<{\centering}p{1.4cm}<{\centering}p{1.2cm}<{\centering}p{1.3cm}<{\centering}p{1.3cm}<{\centering}p{1.1cm}<{\centering}
        }
            \hline
            Model &
		      Accuracy &
            Completeness &
            Relevance &
            Naturalness &
            Proactivity &
            Average \\
		\hline
        Ming-Lite-Omni & 44.63 & 40.58 & 69.28 &  95.65 & 28.78 & 55.78 \\
        Qwen2.5-Omni & 54.03 & 53.68 & 78.00 &  98.28 & 46.25 & 66.05 \\
    	Qwen3-Omni & 60.91 & 58.03 & 83.65 & 99.31 & 42.51 & 68.88 \\
    	Gemini-2.5-Pro & 68.03 & \underline{\textbf{69.07}} & 86.85 & 98.99 & 52.02 & 74.99 \\
        \hline
        \modelflash{} & \underline{\textbf{71.23}} & 65.07 & \underline{\textbf{90.07}} & \underline{\textbf{99.36}} & \underline{\textbf{53.85}} & \underline{\textbf{75.92}} \\
		\hline
	\end{tabular}
\end{table*}

\subsection{Quantitative Results}
\label{sec:B_5_3_performancecomp}

We conduct comprehensive evaluations of \modelflash{} against state-of-the-art MLLMs on over 50 different multimodal benchmarks, as illustrated in Table \ref{table_1_image2text}$\sim$\ref{table_11_streamingmultiturnbench}. Extensive experiments demonstrate that \modelflash{} achieves comparable performance with leading MLLMs.

\noindent\textbf{Vision $\rightarrow$ Text (Understanding)} As shown in Table~\ref{table_1_image2text}, \modelflash{} demonstrates strong and balanced performance across multimodal tasks, standing out among open omni-modal models. It achieves top scores on HallusionBench (66.1) and MMVet (85.6), reflecting robust visual reasoning and low hallucination. On MMStar, it scores 74.9, slightly below GLM-4.6-V (75.9) but ahead of Qwen3-Omni (68.5) and Gemini 2.5 Pro (73.6). Its dominant result on the in-house Wiki Knowledge benchmark (64.6 vs. $\leq$ 48.7 for others) highlights superior factual grounding. In STEM reasoning, \modelflash{} attains the best overall score on MathVerse\textsubscript{Vision} (75.5), where success depends on interpreting visual mathematical content, and leads open models on MMMU (77.1). On OCR tasks, it sets a new high on OCRBench (891), though it is slightly behind GLM-4.6-V on ChartQA and DocVQA. In multi-image understanding, it outperforms Qwen3-Omni on MMT-Bench (70.1 vs. 69.9) and LLaVA-Interleave (63.1 vs. 61.6), but lags on MuirBench.

In video understanding, \modelflash{} achieves state-of-the-art results on MVBench (77.5), surpassing all other open models, and performs competitively on moment localization, dynamic reasoning, and long-form video tasks—trailing only closed-source or larger-scale systems like Gemini 2.5 Pro. Overall, it offers state-of-the-art open performance with particular strengths in visually grounded reasoning, factual alignment, and document understanding.

\noindent\textbf{Text $\rightarrow$ Image (Generation)}. As shown in Table \ref{table_g_imggen}, our experimental results demonstrate that the generation quality of \modelflash{} is on par with state-of-the-art diffusion models. 
Notably, on the Geneval benchmark, our model surpasses all previous methods, demonstrating exceptional controllability. This advantage is particularly pronounced in the "Position" and "Color." sub-categories.
% On the DPG-Bench benchmark, \modelflash{} achieves an overall score of 86.98, outperforming pure image generation models like SD3-Medium (84.08) and leading unified models like OmniGen2 (83.57).
On the DPG-Bench benchmark, \modelflash{} achieves an overall score of 86.98, closely approaching the state-of-the-art performance of 88.32 achieved by Qwen-Image, while substantially outperforming pure diffusion models such as SD3-Medium (84.08).

\noindent\textbf{Image $\rightarrow$ Image (Editing)}. As shown in Table \ref{table_i2i_gedit}, 
\modelflash{} demonstrates impressive image editing performance, surpassing all other models. 
% Specifically, this superiority is reflected in its strong scores for both Semantic Consistency (G\_SC) and Perceptual Quality (G\_PQ), which align with its high Overall Score (G\_O). 
Specifically, \modelflash{} supports editing instructions in Chinese, achieving performance comparable to that with English instructions. 
% Compared to Qwen-Image-Edit which utilizes a 20B DiT head, \modelflash{} achieves comparable semantic consistency and perceptual quality with a much more efficient 2B DiT head—only one-tenth the parameters. This efficiency also translates to remarkable inference speeds, typically between 1-2 seconds per generation.
Compared to Z-Image-Edit, which is specifically trained for editing tasks, \modelflash{} achieves comparable semantic consistency while delivering superior perceptual quality and higher overall scores under joint training for image generation and editing.
% 与 专为编辑任务进行训练的 Z-Image-Edit相比，\modelflash{} 在联合图像生成和编辑训练下 achieves comparable semantic consistency and 更优的 perceptual quality 与 overall score， 
% Compared with Z-Image-Edit, which is specifically trained for editing tasks, \modelflash{}—trained jointly for image generation and editing—achieves comparable semantic consistency while delivering superior perceptual quality and a higher overall score.
% Compared to Z-Image-Edit, which is specifically trained for editing tasks, \modelflash{} achieves comparable semantic consistency while delivering superior perceptual quality and higher overall scores under joint training for image generation and editing.

\noindent\textbf{Image $\rightarrow$ Image (Segmentation)}. As shown in Table~\ref{table_i2i_seg}, \modelflash{} is capable of performing segmentation tasks, achieving performance comparable to that of specialized models designed explicitly for this purpose. Compared to other unified MLLMs, \modelflash{} demonstrates a significant advantage in segmentation. 
For instance, Qwen-Image-Edit often struggles to accurately localize the target object, while Nano-banana frequently misinterprets user intent during inference.
In contrast, \modelflash{} exhibits superior robustness and a more accurate understanding of spatial and semantic instructions.

% ***--***--***

\noindent\textbf{Audio $\rightarrow$ Text (Understanding)}. As shown in Table \ref{tab:main_context_asr_perf}, Ming-Flash-Omni achieves the best overall average (Avg) on ContextASR-Bench, indicating robust context utilization. On several individual subsets/metrics, Qwen3-ASR\citep{shi2026qwen3asrtechnicalreport} and Qwen3-Omni-30B-A3B-Instruct\citep{xu2025qwen3omnitechnicalreport} perform better, suggesting complementary strengths. It also exhibits highly competitive performance across various ASR benchmarks, with notable strengths in dialect recognition (Table \ref{table_8_audioasr}). In audio question answering, \modelflash{} surpasses all open-source audio-centric and other Omni models, with the exception of Qwen3-Omni-Flash-Instruct\citep{xu2025qwen3omnitechnicalreport} (Table \ref{table_9_audioqa}). Taken together, these findings demonstrate the robust and versatile audio understanding capabilities of \modelflash{}.

% ***--***--***

\noindent\textbf{Text $\rightarrow$ Audio (Generation)}. As shown in Table \ref{table_10_audiotts}, leveraging advancements in speech representation and model architecture, \modelflash{} achieves SOTA performance among open-source models of a comparable size on the test-zh subset of the SEED-TTS-Eval benchmark\citep{anastassiou2024seed}. For controllable tasks, \modelflash{} achieves an accuracy (ACC) of 83.25\% on the Sichuanese dialect, outperforming CosyVoice3 (as shown in Table \ref{tab::audiotts_dialect_sichuanese}). On the podcast task, \modelflash{} sets the new state-of-the-art (SOTA) among open-source models for CER and UTMOS on the ZipVoice-Dia(zh) benchmark (see Table \ref{tab:podcast_results}).

\noindent\textbf{Video + Audio $\rightarrow$ Text (Video Streaming Conversation)}.
As shown in Table \ref{table_11_streamingmultiturnbench}, benefiting from the introduction of high-quality and diverse streaming video multiturn data, \modelflash{} has achieved significant improvements in all dimensions compared to \modellite{}. \modelflash{} also outperforms Qwen3-Omni~\citep{xu2025qwen3omnitechnicalreport} and Gemini 2.5-Pro~\citep{comanici2025gemini25} in the dimensions of accuracy, completeness, and relevance, providing better experience in streaming video conversation scenarios.

\begin{figure*}[bthp]
    \centering
    \includegraphics[width=0.9\linewidth]{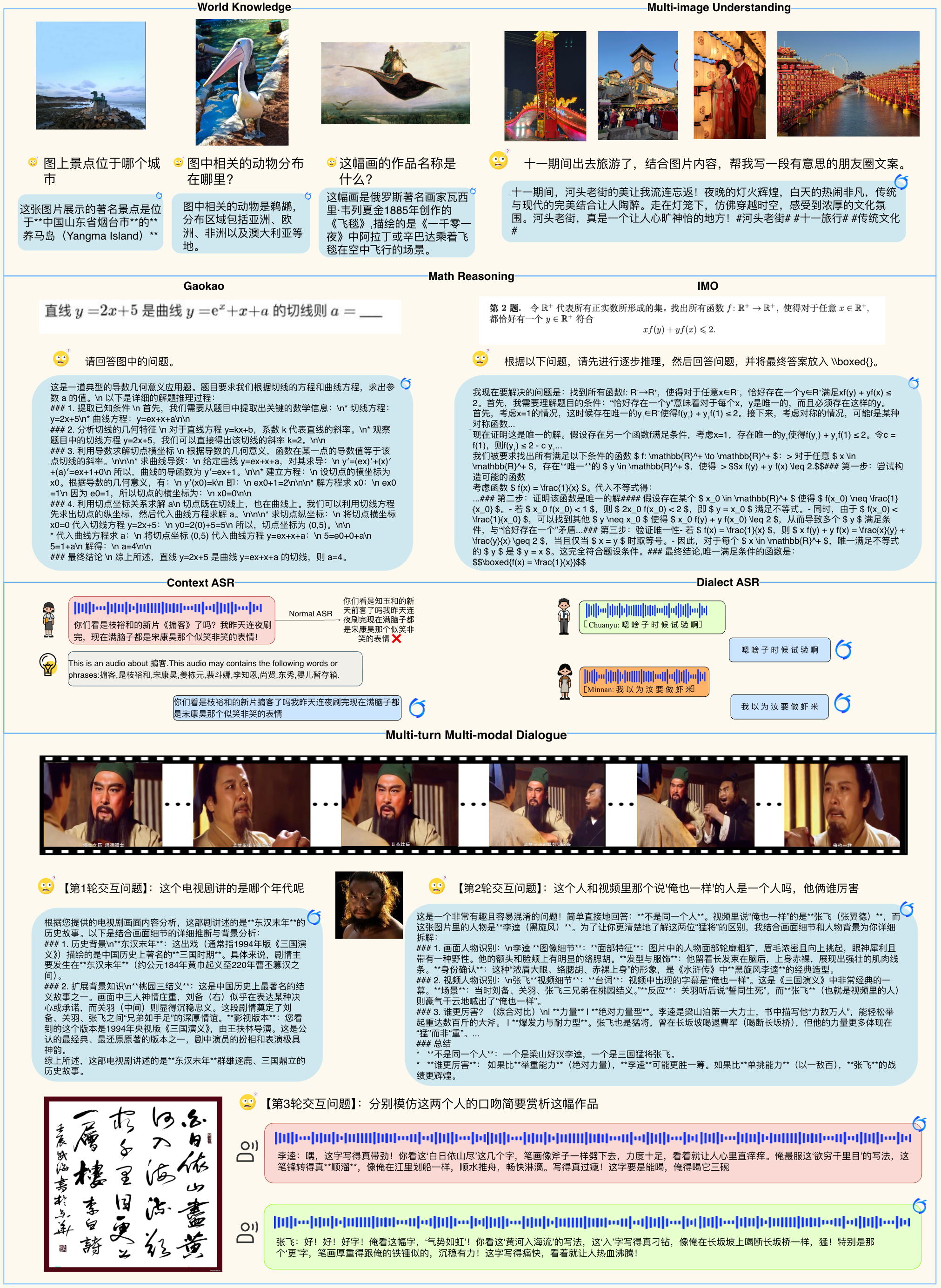}
    \caption{
        Qualitative results of \modelflash{} across diverse understanding tasks, including world knowledge grounding, multi-image reasoning, mathematical problem solving, contextual and dialect-aware ASR, and multi-turn multimodal dialogue.
    }
    \label{visualization}
\end{figure*}

\begin{figure*}[tbhp]
    \centering
    \includegraphics[width=0.93\linewidth]{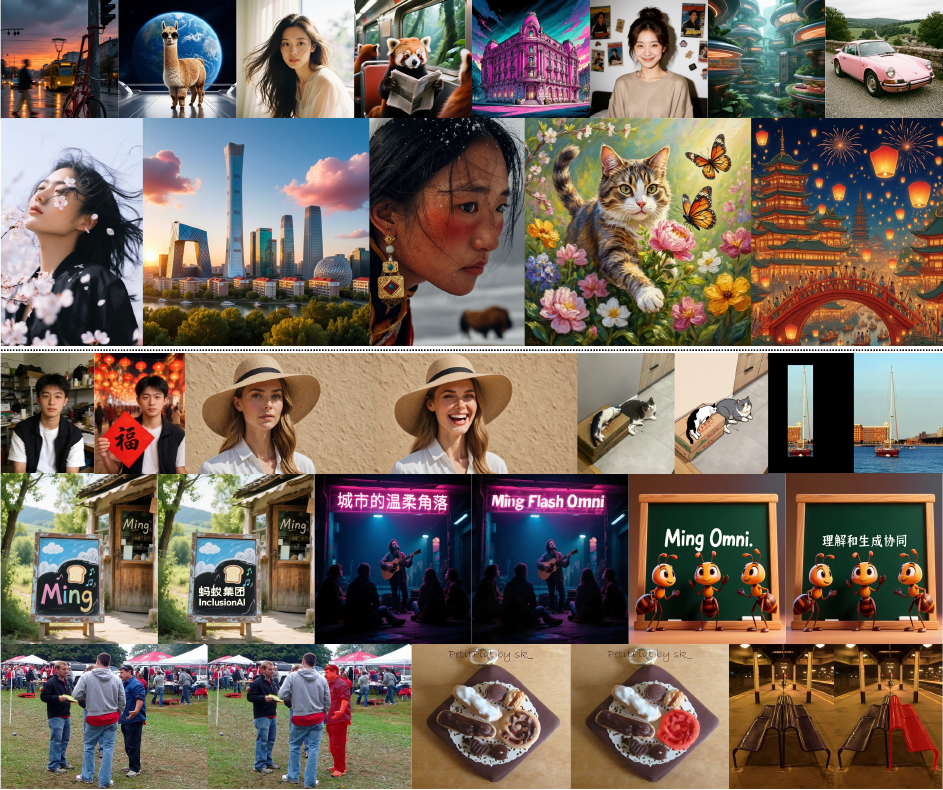}
    \caption{
        Visualization results of \modelflash{} on {Text/Image $\rightarrow$ Image} tasks, including image generation task, image editing task, and image segmentation task.
    }
    \label{part-5-fig-4-vis-gen}
\end{figure*}

\begin{figure*}[bthp]
    \centering
    \includegraphics[width=0.93\linewidth]{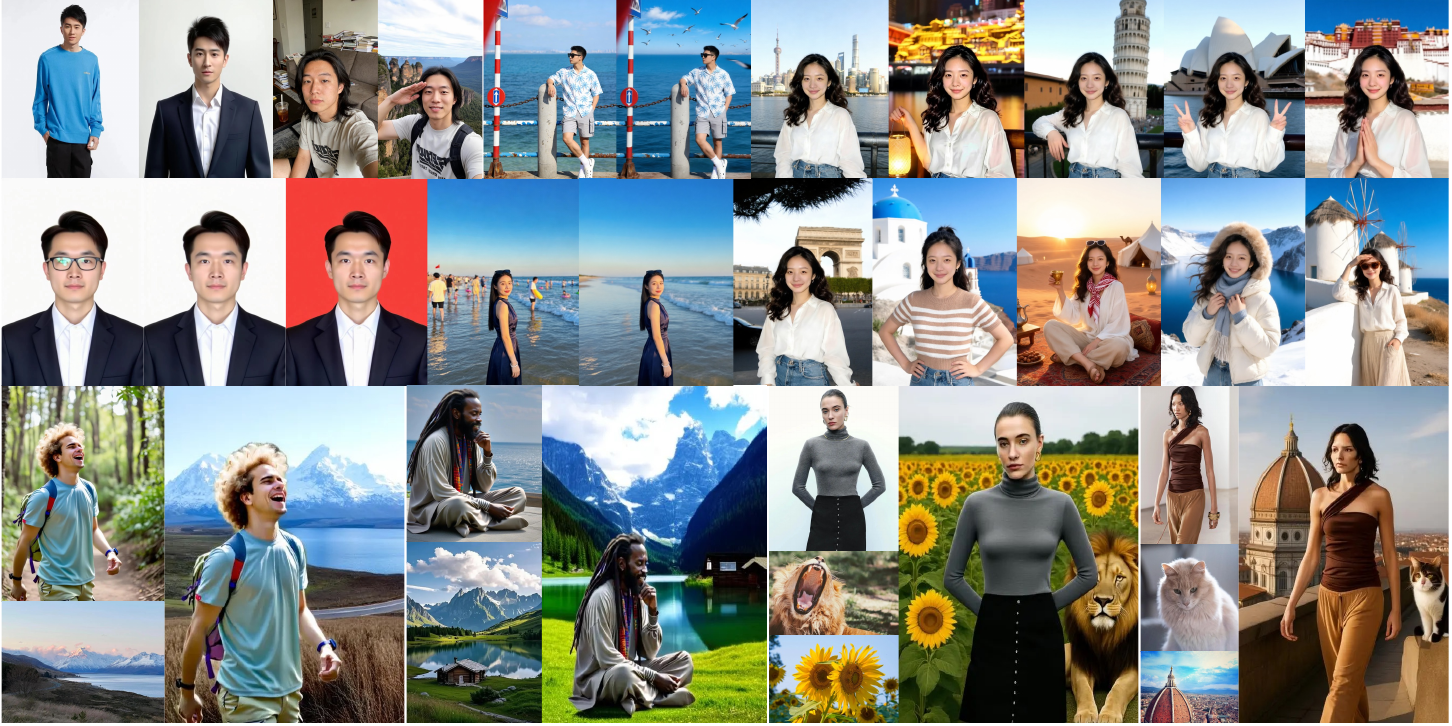}
    \caption{
        Visualization results of \modelflash{} on {Image $\rightarrow$ Image} tasks, including ID photo generation, ID photo editing, background replacement, and multi-image editing.
% 证件照生成 证件照编辑 背景变化 多图编辑
    }
    \label{part-5-fig-4-vis-gen-multi}
\end{figure*}

\subsection{Visualization Results}
\label{sec:B_5_3_visualization}

In this section, we present several qualitative examples to illustrate the multifaceted capabilities of \modelflash{}. As shown in the figure, \modelflash{} exhibits strong visual understanding across multiple dimensions. In world knowledge grounding, it accurately identifies landmarks, animal species, and famous artworks from images—demonstrating its ability to integrate visual perception with rich encyclopedic knowledge. In mathematical reasoning, it solves complex problems ranging from Chinese college entrance exam (Gaokao) questions to International Mathematical Olympiad (IMO)-level challenges, providing clear, step-by-step derivations that reflect deep logical understanding. For multi-image understanding, it generates creative and coherent narratives grounded in multiple input images, showcasing its capacity for cross-image reasoning and contextual synthesis. Turning to speech recognition, ~\modelflash{} achieves strong performance on contextual ASR tasks. By leveraging contextual information, it effectively resolves many challenging cases where conventional ASR systems tend to fail—such as ambiguous homophones, domain-specific terminology, or noisy conversational speech. Moreover, this version also supports multiple Chinese dialects, significantly broadening its applicability in real-world multilingual and regional speech scenarios. Notably, \modelflash{} also supports multi-turn multimodal dialogues with seamless switching between modalities and tasks, enabling fluid interactions that combine vision, speech, and text in a unified conversational framework. Lastly, we visualize the capabilities of \textit{Text/Image 
→Image} generation tasks in Figure~\ref{part-5-fig-4-vis-gen} and Figure~\ref{part-5-fig-4-vis-gen-multi}, covering a wide range of applications including image generation, image editing, image segmentation, multi-image editing, ID photo generation, ID photo editing, and background replacement.
As can be seen, ~\modelflash{} not only supports a broader set of generative capabilities but also achieves higher output quality and greater controllability compared to previous versions.

\section{Conclusion}
\label{sec:Conclusion}
% 
% 不需要highlight sparse，展开upgraded 展开一下是，哪些方面，概括一点，和方法data部分相关
% SOTA performance on a range of-> 4个 多图，视频，生成，分割，contextASR和方言，在xxx个榜单取得了xxx，a range of xxx xx modal include xx

In this paper, we present \modelflash{}, built upon Ling-Flash-2.0 with 100 billion parameters, where only 6.1B parameters are activated per token.  \modelflash{} demonstrates advanced multimodal perception and generation capabilities with improved computational efficiency while scaling model capacity. It achieves SOTA performance across a broad spectrum of tasks, including multi-image and video processing, image generation, generative segmentation, Contextual Automatic Speech Recognition (ContextASR), and multi-dialect recognition, outperforming omni models of comparable scale. We believe the open-sourcing of our models and code will facilitate the development of AGI by advancing multimodal intelligence research and enabling broader real-world applications.

\newpage
\section{Contributors}
\label{sec:contri}

\large{Authors are listed \textbf{alphabetically by the first name}.}

\large{
\begin{multicols}{3}
\raggedcolumns
Ant Inclusion AI \\
Baihui Li\\
Bowen Ma\\
Cheng Zou\\
ChengKun Du \\
Canxiang Yan \\
Chunxiang Jin \\ 
Chunjie Shen \\
Chenyu Lian \\
Chengxiang Fan \\
Dandan Zheng \\
Fudong Wang\\
Furong Xu\\
Guangming Yao\\
Haohao Liu \\
Han Peng \\
Jun Zhou \\
Junluan Xia \\
Jingdong Chen \\
Jianing Li \\
Jianxin Sun \\
Jianjiang Zhu \\
Jianping Jiang \\
Jinpeng Ou \\
Jun Peng\\
Jin Peng\\
Kaixiang Ji\\
Li Tang\\
Libin Wang\\
Lixiang Ru\\
Longhua Tan \\
Lu Ma \\
Lan Wang \\
Mochen Bai\\
Minghong Cai\\
Mingxue Yang\\
Ning Gao \\
Qingpei Guo \\
Qinglong Zhang \\
Qiang Xu \\
Qin Zhao\\
Rui Liu \\
Ruijie Xiong \\
Ruobing Zheng \\
Sirui Gao \\
Shaoxiong Lin \\
Tao Zhang \\
Tianqi Li \\
Tinghao Liu \\
Tongli Wang \\
Taoye Huang \\
Weilong Chai \\
Xiaomei Wang \\
Xiaolong Wang \\
Xiaojian Liu \\
Xiao Lu \\
Xiaoyu Li \\
Xingning Dong\\
Xuzheng Yu\\
Xuezhi Wang\\
Yi Yuan \\
Yuting Gao\\
Yuting Xiao\\
Yunxiao Sun\\
Yipeng Chen\\
Yifan Mao \\
Yifei Wu\\ 
Yingying Zhang\\
Yongjie Lyu\\ 
YuQian Li\\
Ziping Ma\\
Zhiqiang Fang\\
Zhihao Qiu\\
Ziyuan Huang \\
Zizheng Yang\\
Zhengyu He

\end{multicols}}

\clearpage

\bibliographystyle{assets/plainnat}
\bibliography{paper}

\newpage
\beginappendix

\section{Public Benchmarks}
\label{sec:app_open_benchmark}

% \subsection{Public Benchmarks}
% \label{sec:app_1_public}
% \label{sec:B_5_1_public}

\noindent\textbf{Image $\rightarrow$ Text (Understanding)}. Our evaluation of the image-to-text understanding capabilities primarily encompasses the following six tasks: 1) general image understanding capabilities evaluated on MMStar~\citep{chen2024we}, AI2D~\citep{kembhavi2016diagram}, HallusionBench~\citep{Guan_2024_hallusionbench}, MMBench~\citep{liu2024mmbench}, MMVet~\citep{yu2023mm}. 2) STEM and reasoning capabilities are evaluated on MMMU\textsubscript{val}~\citep{yue2024mmmu}, MathVista\textsubscript{mini}~\citep{lu2024mathvista}, MathVerse\textsubscript{Vision}~\citep{zhang2024mathverse}, MathVision\citep{wang2024measuring}, LogicVista~\citep{xiao2024logicvista}. 3) OCR capabilities evaluated on ChartQA~\citep{masry-etal-2022-chartqa}, DocVQA~\citep{docvqa} and OCRBench~\citep{Liu_2024}. 4) multi-image capabilities evaluated on MMTBench~\citep{ying2024mmtbench}, MuirBench~\citep{wang2024muirbench}, and LLaVA-interleave Bench~\citep{li2024llava_interleave_bench}. 

\noindent\textbf{Video $\rightarrow$ Text (Understanding)}. Our evaluation of the video-to-text understanding capabilities contains the following four benchmarks: MVBench~\citep{li2024mvbench}, VideoMME~\citep{fu2024video}, LongVideoBench~\citep{wu2024longvideobench}, MLVU~\citep{zhou2024mlvu}, PerceptionTest~\citep{patraucean2023perception}, CharadesSTA~\citep{song2024temporal} and TOMATO~\citep{shangguan2025tomatoassessingvisualtemporal}.

\noindent\textbf{Text $\rightarrow$ Image (Generation)}. We incorporate text-to-image generation capabilities to enable our MLLM with unified perception-generation abilities, which are evaluated on GenEval~\citep{ghosh2024geneval} and DPG-Bench~\citep{hu2024ella}.

\noindent\textbf{Image $\rightarrow$ Image (Editing)}. Our evaluation of image-to-image editing capabilities is conducted on the GEdit-Bench benchmark~\citep{liu2025step1x}.

\noindent\textbf{Image $\rightarrow$ Image (Segmentation)}. 
We evaluate the segmentation capability of our MLLM on the standard referring expression segmentation (RES) benchmarks RefCOCO/+~\citep{refcoco} and RefCOCOg~\citep{refcocog}.

\noindent\textbf{Audio $\rightarrow$ Text (Understanding)}. Our evaluation of the audio-to-text understanding capabilities mainly includes the following three tasks: 1) Fundamental audio understanding capabilities evaluated on a broad range of public benchmarks, including public Chinese benchmarks like Aishell1~\citep{AISHELL1} and Wenetspeech~\citep{WenetSpeech}, and public English benchmarks like Librispeech~\citep{Librispeech} and Voxpopuli~\citep{wang2021voxpopuli}. And 2) audio question-answering capabilities evaluated on various benchmarks across five specific tasks, such as AlpacaEval and CommonEval from VoiceBench~\citep{chen2024voicebenchbenchmarkingllmbasedvoice} for open-ended QA tasks, and SD-QA for knowledge-based QA tasks. Finally 3) evaluates the model's ability to utilize context on ContextASR-Bench\citep{wang2025contextasrbenchmassivecontextualspeech}.

\noindent\textbf{Text $\rightarrow$ Audio (Generation)}. We incorporate text-to-audio generation capabilities to enable our MLLM with unified audio perception-generation abilities, which are evaluated on Seed-TTS-Eval~\citep{Seed_TTS}. For the controllable tasks, we use Wenetspeech-chuan(WSC)~\citep{dai2025wenetspeech} to evaluate the performance of dialect generation. For the podcast tasks, we utilize ZipVoice~\citep{zhu2025zipvoice} for evaluation.

\section{The prompt used in data generation}
This section presents the prompt used for data generation in Sec.~\ref{sec:data}. 
% In this section, we present the prompt used in our method for data generation.
 
% \textbf{Prompt for portrait preservation data}

\begin{figure}[htbp]
\centering
\begin{tcolorbox}[colback=black!5!white,colframe=black!75!black,title=Prompt for portrait preservation data]

\begin{lstlisting}[language=Python, breakindent=0pt]
Based on the person's age and gender in the image, generate a detailed description of a realistic life scene photo.
Requirements:
1. Must include: clothing, location, action, style.
2. Strictly within 50 words.
3. Change the scene (real scenes: indoor, outdoor, residential areas, homes, companies, kitchens, parks, etc.) and character costumes to ensure that even similar images have different descriptions.
4. Clothing must match age and gender, vary descriptions.
5. Scenario details must include at least 3 realistic elements, be vivid and lifelike.
Output only the description in English, no additional text.
Example: A young woman with medium-length dark hair tied back in a neat ponytail stands outdoors near a park bench, dressed in a crisp white button-up shirt under a fitted black blazer. She wears a subtle headband, her brow knitted as she runs her fingers through her hair, her lips slightly downturned and eyes glistening with unshed tears. The scene captures her from a distant angle, with trees and a blurred pathway in the soft-focus background, suggesting contemplation or distress amidst nature.
 
\end{lstlisting}
\label{prompt1}
\end{tcolorbox}
\end{figure}

\begin{figure}[htbp!]
\centering
\begin{tcolorbox}[colback=black!5!white,colframe=black!75!black,title=Prompt for text generation data]

\begin{lstlisting}[language=Python, breakindent=0pt]
Describe according to the following requirements:
1. Randomly select a theme from {theme}, and generate text content in Chinese, English, and numbers, with 3-5 characters.
2. Use your imagination; the theme is not fixed. Keep the description under 100 characters.
3. Generate a suitable image description based on the text content.
4. Refer to the {text style} and choose appropriate font, color, and layout.

Output format:
Description, Text: "content", Font style, Font color, and Image layout

Examples:
In the image, there is a gym scene, Text: "Power 3.0", font style is broken art font with cracks and shattering effects, font color is metallic gray with dark red gradient, text layout is centered-right, background includes dumbbells, treadmills, and reflective glass walls.

\end{lstlisting}
\label{prompt2}
\end{tcolorbox}
\end{figure}
% In the image, a calm ocean with a vibrant sunset, Text: "Do not be sad", font style is dynamic flame art font, color is orange-red gradient, text centered, flame effects radiate from the letters outward, full of explosive energy and visual impact.

% \textbf{Prompt for text generation data}

% \textbf{Prompt for video multi-turn conversation data}

\begin{figure}[htbp]
\centering
\begin{tcolorbox}[colback=black!5!white,colframe=black!75!black,title=Prompt for video multi-turn conversation data]

% You are an expert video analyst. Your task is to analyze the provided video and output a structured JSON object containing your analysis. You must adhere strictly to the format and rules described below.

\begin{lstlisting}[language=Python, breakindent=0pt]
  # Role: Expert Video Analyst
  You are an expert video analyst. Your task is to analyze the provided video and output a structured JSON object containing your analysis. You must adhere strictly to the format and rules described below.
  
  # Instructions:
  Analyze the video and generate a single JSON object with the following keys. Your response should ONLY be the JSON object, enclosed in a single markdown code block (```json ... ```). Do not include any other text, explanations, or introductory phrases.

  # JSON Fields Definition:
  1.  `"category"`: (String) Select ONLY ONE category from the following list that best describes the main subject of the video. 
  If the video category is not among the listed below, output  "Others".
  * **List of 34 Categories**: [
       "Others", "LifeRecord-TravelLog", "LifeRecord-DailyLife", "LifeRecord-HouseTour", "LifeRecord-Reaction", "LifeRecord-AnimalPet", "LifeRecord-Cooking", "LifeRecord-Fashion", "LifeRecord-Workout", "Education-Lecture", "Education-Finance", "Education-Multilingual", "Education-Handifraft", "Education-Science", "Education-Art", "Education-OnlineTutorial", "TVShow-TVSeries", "TVShow-News", "TVShow-TalkShow", "TVShow-Celebration", "TVShow-CommentaryProgram", "Competition-Football", "Competition-Athletics", "Competition-Basketball", "Competition-Snooker", "Competition-Boxing", "Competition-Car", "VideoGames-Sandbox", "VideoGames-OpenWorld", "Documentary-Nature", "Documentary-Science", "Documentary-Culture", "Documentary-Kids", "Movie-Comedy", "Movie-Adventure"]

  2.  `"caption"`: (String) A concise but descriptive summary of the video's content in English, not exceeding 30 words. Describe what is happening, who is involved, and the main subject.

  3.  `"content_score"`: (Integer) An integer from 1 to 10. This score evaluates the video's potential as a prompt for a synthetic AI-user conversation.
  * **Score 1-3 (Low)**: The video content is sparse, repetitive, or features a single person's monologue with little interaction or environmental detail. It's difficult to ask questions or start a conversation based on it.
  * **Score 4-7 (Medium)**: The video has some interesting elements but may lack depth or variety. It can support a few questions but may not lead to an extended, rich conversation.
  * **Score 8-10 (High)**: The video is rich in content, detail, and action. It shows a process, an interaction, or a complex scene that naturally invites questions and allows for a deep, extended conversation (e.g., a detailed cooking tutorial, a complex assembly process, a travel vlog with multiple activities).

  4.  `"fov"`: (Integer) Field of View.
  * **1**: The video is shot from a first-person perspective (FPV), where the camera acts as the viewer's eyes.
  * **0**: The video is shot from a third-person or static perspective.

\end{lstlisting}

\label{prompt3}
\end{tcolorbox}
\end{figure}

\begin{figure}[t!]
\centering
\begin{tcolorbox}[colback=black!5!white,colframe=black!75!black,title=]

\begin{lstlisting}[language=Python, breakindent=0pt]
  5.  `"task_complexity"`: (Integer) An integer from 1 to 10, representing the complexity of the primary task shown in the video. If no specific task is shown, rate the complexity of the main activity.
  * **Score 1-3 (Low)**: Simple, everyday actions that require minimal skill (e.g., opening a bottle, pouring water, petting a cat).
  * **Score 4-7 (Medium)**: Tasks that require some skill, knowledge, or multiple steps (e.g., cooking a simple dish, assembling IKEA furniture, basic makeup application).
  * **Score 8-10 (High)**: Highly complex, specialized, or professional tasks that require significant expertise, precision, or effort (e.g., performing surgery, building a car engine, professional programming, playing a complex musical piece).

  # Required Output Format:
      ```json
      {
        "category": "...",
        "caption": "...",
        "content_score": ...,
        "fov": ...,
        "task_complexity": ...
      }
  ```
\end{lstlisting}
\label{prompt4}
\end{tcolorbox}
\end{figure}

\end{document}